\documentclass[lettersize,journal]{IEEEtran}
\usepackage{amsmath,amsfonts}
\usepackage{amsthm,amsmath,amssymb}
\usepackage{mathrsfs}
\usepackage{algorithmic}
\usepackage{algorithm}
\usepackage{array}
\usepackage[caption=false,font=normalsize,labelfont=sf,textfont=sf]{subfig}
\usepackage{textcomp}
\usepackage{stfloats}
\usepackage{url}
\usepackage{bm}
\usepackage{verbatim}
\usepackage{graphicx}
\usepackage{cite}
\usepackage{amssymb}
\usepackage{booktabs,makecell,multirow}
\usepackage{threeparttable}
\usepackage{color}
\makeatletter
\renewcommand{\maketag@@@}[1]{\hbox{\m@th\normalsize\normalfont#1}}%
\makeatother
\begin{document}

\title{CS-Net: Contribution-based Sampling Network for Point Cloud Simplification}

\author{Tian Guo, Chen Chen, Hui Yuan,~\IEEEmembership{Senior Member,~IEEE,} Xiaolong Mao, Raouf Hamzaoui,~\IEEEmembership{Senior Member,~IEEE,} and Junhui Hou,~\IEEEmembership{Senior Member,~IEEE}
\thanks{This work was supported in part by the National Natural Science Foundation of China under Grants 62222110, 62172259, and 62311530104, the High-end Foreign Experts Recruitment Plan of Chinese Ministry of Science and Technology under Grant G2023150003L, the Taishan Scholar Project of Shandong Province (tsqn202103001), the Natural Science Foundation of Shandong Province under Grant ZR2022ZD38, and the OPPO Research Fund.}
\thanks{Tian Guo, Chen Chen, and Hui Yuan are with the School of Control Science and Engineering, Shandong University, Ji'nan, 250061, China (e-mail: guotiansdu@mail.sdu.edu.cn; chenc\_nj@mail.sdu.edu.cn; huiyuan@sdu.edu.cn).}
\thanks{Xiaolong Mao is with the School of Software, Shandong University, Ji'nan, 250061, China (e-mail: xiaolongmao@mail.sdu.edu.cn).}
\thanks{Raouf Hamzaoui is with the School of Engineering and Sustainable Development, De Montfort University, LE1 9BH Leicester, UK. (e-mail: rhamzaoui@dmu.ac.uk).}
\thanks{Junhui Hou is with the Department of Computer Science, City University of Hong Kong, Hong Kong, China (e-mail: jh.hou@cityu.edu.hk).}
\thanks{Hui Yuan is the corresponding author.}
}

\markboth{Journal of \LaTeX\ Class Files,~Vol.~14, No.~8, August~2021}%
{Shell \MakeLowercase{\textit{et al.}}: CS-Net: Contribution-based Sampling Network for Point Cloud Simplification}


\maketitle

\begin{abstract}
Point cloud sampling plays a crucial role in reducing computation costs and storage requirements for various vision tasks. Traditional sampling methods, such as farthest point sampling, lack task-specific information and, as a result, cannot guarantee optimal performance in specific applications. Learning-based methods train a network to sample the point cloud for the targeted downstream task. However, they do not guarantee that the sampled points are the most relevant ones. Moreover, they may result in duplicate sampled points, which requires completion of the sampled point cloud through post-processing techniques. To address these limitations, we propose a contribution-based sampling network (CS-Net), where the sampling operation is formulated as a Top-\(k\) operation. To ensure that the network can be trained in an end-to-end way using gradient descent algorithms, we use a differentiable approximation to the Top-\(k\) operation via entropy regularization of an optimal transport problem. Our network consists of a feature embedding module, a cascade attention module, and a contribution scoring module. The feature embedding module includes a specifically designed spatial pooling layer to reduce parameters while preserving important features. The cascade attention module combines the outputs of three skip connected offset attention layers to emphasize the attractive features and suppress less important ones. The contribution scoring module generates a contribution score for each point and guides the sampling process to prioritize the most important ones. Experiments on the ModelNet40 and PU147 showed that CS-Net achieved state-of-the-art performance in two semantic-based downstream tasks (classification and registration) and two reconstruction-based tasks (compression and surface reconstruction). CS-Net also achieved high average precision for objection detection on the KITTI LiDAR point cloud dataset, demonstrating its effectiveness in three-dimensional object detection.
\end{abstract}


\section{Introduction}
\IEEEPARstart{A} three-dimensional (3D) point cloud is a set of unordered 3D points that are characterized by their geometry coordinates and attribute information \cite{refb,refc,refa}. Given the recent strides in 3D sensing technology \cite{refd,refd2}, 3D point clouds are attracting increasing attention across diverse fields, such as 3D object classification \cite{ref1,ref2,refq,refw} and 3D scene reconstruction \cite{ref3,ref4,refe,refr}. However, processing point clouds, especially large-scale ones, imposes significant storage requirements and computational demands, which limits their applicability. Point cloud sampling reduces storage and transmission requirements without introducing significant accuracy loss. 

Point cloud sampling is typically divided into generative (or soft) sampling and selective (or hard) sampling methods based on whether the sampled point cloud is a subset of the input point cloud. Generative sampling methods generate points that may not belong to the input point cloud. This may affect the similarity between the sampled point cloud and the original one. Selective sampling methods directly choose points from the input point cloud based on matrix multiplication. However, this approach may result in the selection of duplicate points. Traditional selective methods, such as farthest point sampling (FPS) \cite{ref5} and Poisson disk sampling \cite{ref6}, treat all points equally and do not take downstream tasks into account. Recently, learning-based sampling networks \cite{ref9,ref10,ref11} have been proposed to generate point clouds optimized for the downstream task. However, these methods do not ensure that the sampled points are the most relevant. Moreover, the network may output duplicate sampled points, necessitating an additional step to complete the sampled point cloud.

In this paper, we introduce CS-Net, a contribution-based sampling network. It includes a feature embedding module designed to extract both local and global features and an attention mechanism to emphasize critical features. Since the attention mechanism may focus on different features in each layer, the outputs of each layer are combined to retain the extracted features. Unlike existing learning-based sampling methods, including our previous work \cite{ref33}, CS-Net proposes a contribution scoring module that generates a contribution score for each point and guides the sampling process to prioritize the most important ones. 

Our contributions are as follows. 
\begin{itemize}
\item{We formulate selective point cloud sampling as a Top-\(k\) operation where the input points are ranked according to their importance.}
\item{We propose a point cloud sampling network, namely CS-Net. Our network consists of three novel modules: a feature embedding module, a cascade attention module, and a contribution scoring module.}
\item{We ensure that our network can be trained end-to-end by modeling the Top-\(k\) operation as the optimal solution to an optimal transmission problem regularized with an entropic penalty.}
\item{We guarantee that there are no duplicate points in the sampled point cloud without the need of post-processing techniques. }
\item{We introduce Earth Mover's Distance (EMD) into the loss function to ensure the sampled point cloud preserves the original shape by effectively regulating the distribution of points.}
\item{Experimental results show that the proposed network outperforms the state-of-the-art in four downstream tasks.}
\end{itemize}

The remainder of this paper is organized as follows. In Section II, we briefly review related work. In Section III, we provide the theoretical motivation of our method and formulate the problem mathematically. In Section IV, we present and analyze the proposed network. Experimental results and conclusions are given in Section V and VI, respectively.
\vspace{-6pt}
\section{Related work}
Point cloud sampling can be classified into generative sampling and selective sampling. Generative sampling generates points directly based on learned features, resulting in the generated point cloud that is not a subset of the input point cloud. While selective sampling selects points from the input point cloud based on predefined rules or extracted features, ensuring that the sampled point cloud is a subset of the input point cloud.
\vspace{-8pt}
\subsection{Generative Point Cloud Sampling Methods}
Dovrat \textit{et al.} \cite{ref9} proposed S-Net, the first sampling network that is trained for a given downstream task. S-Net generates an output point cloud by using a fully connected layer. Since the generated point cloud is not necessarily a subset of the input point cloud, the generated points are matched to the input points in a post-processing step. Consequently, the sampling process is not differentiable. Moreover, the matching process may lead to duplicate points. To address the non-differentiability issue of S-Net, Lang, Manor, and Avidan \cite{ref10} introduced a differentiable relaxation to the matching step. While this relaxation makes the sampling process differentiable, it cannot avoid duplicate points. Lin \textit{et al.} \cite{ref11} observed the existence of overlapped neighborhoods in the projecting operation of SampleNet. They proposed an improved local adjustment module that better preserves the local details and achieves more accurate  classification compared with SampleNet. Lin \textit{et al.} \cite{ref12} proposed DA-Net, a network introducing a density-adaptive downsampling module. This module can adaptively adjust the sampling rate across various regions of the point cloud by taking the estimated local density into account. Wang \textit{et al.} \cite{ref13} devised a point sampling transformer network by combining S-Net and transformer. Benefiting from transformer, this method can generate a noise-insensitive point cloud. Tian \textit{et al.} \cite{ref14} introduced a universal point cloud sampling network capable of extracting representative points without the need for task-specific fine-tuning. However, this method retains the matching operation proposed by S-Net, which cannot be trained. While various approaches have been proposed to enhance the accuracy of the sampled point clouds in approximating the input point clouds, one common problem is that the generated point cloud is not necessarily a subset of the input point cloud. Hence, unlike traditional methods such as FPS, these methods cannot always preserve the shape of the original point cloud, potentially resulting in poorer subjective quality.
\vspace{-10pt}
\subsection{Selective Point Cloud Sampling Methods}

\textit{1) Traditional Selective Sampling Methods.}
The most commonly used traditional methods are random sampling (RS) \cite{ref8}, FPS \cite{ref5}, and Poisson-disk sampling \cite{ref6}. In random sampling, a subset of points is randomly selected from the input point cloud. This method is simple and efficient but suffers from some drawbacks such as uneven point distribution and loss of important semantic features. FPS iteratively selects points that are farthest from the previously selected points. It can generate a subset of points that are evenly distributed across the point cloud. However, the computational complexity of FPS is high, especially when dealing with large-scale point clouds. Poisson-disk sampling generates points by rejecting any new point that is too close to previously generated points. This method can provide uniformly distributed points in a given space with similar computational complexity to FPS.

\textit{2) Learning-based Selective Sampling Methods.}
Learning-based selective sampling methods directly select points from the input point cloud based on specific rules or extracted features. One significant issue with these methods is that the process of selecting the points is discrete and hence not differentiable. Several strategies have been proposed to tackle this problem. Qian \textit{et al.} \cite{ref15} introduced MOPS-Net, an end-to-end deep neural network for task-oriented point cloud sampling. They formulate the sampling problem as a constrained and differentiable matrix optimization problem and design a deep neural network to mimic matrix optimization. Building on MOPS-Net, Yang \textit{et al.} \cite{ref16} proposed an attention-sampling network (AS-Net) that uses an attention module to capture important features. They also provided a constraint matching module to ensure that the sampled point cloud is a subset of the input point cloud. However, this strategy can only be applied during testing. During the training procedure, soft sampling, instead of selective sampling \cite{ref15,ref16}, has to be used. Yang \textit{et al.} \cite{ref17} introduced a Gumbel subset sampling (GSS) method that addresses the challenge by adding Gumbel noise into the binary matrix to solve the trainable problem. They also used a channel shuffle module that enables the interaction of features from different channels without the need for additional parameters. Sun \textit{et al.} \cite{ref18} proposed a sampling network that solves the non-differentiable problem by a gradient estimation strategy. However, the selective sampling method based on matrix multiplication introduces duplicate points in the output point set. Nezhadarya \textit{et al.} \cite{ref19} proposed a deterministic, adaptive, hierarchical sampling method that samples the points according to their importance to the application task. After filtering the input point cloud using a convolutional neural network to derive a set of features, the method selects the subset of points with the highest feature value along each dimension of the feature vectors. These steps are repeated until the desired number of outpoint points is reached. However, the method involves complex operations, such as nearest-neighbor resizing at each iteration. Potamias \textit{et al.} \cite{refd1} proposed a fast point cloud simplification method that overcomes the inefficiencies of traditional greedy methods by using three learnable modules to simplify large-scale point clouds in real-time while preserving salient features and global structural appearance. 

To address the above challenges, we generate a quantitative score for each point, indicating its importance to the downstream task. Points with higher scores are retained in the sampled point cloud using differentiable optimization. Therefore, CS-Net enables end-to-end training of hard sampling and fundamentally resolves the problem of duplicated points in the sampled point set. Moreover, it can sample the input point cloud at any sampling ratio. 

\section{Problem Formulation}
The purpose of selective point cloud sampling is to select the most important points from an input point cloud. If the importance of each point can be quantified by a score, these points can be selected using a Top-\(k\) operator \cite{ref21}. The Top-\(k\) operator finds the \(k\) largest elements in a set. The Top-\(k\) operator maps a set of inputs $x_1, \ldots, x_n$ to an index vector \(\boldsymbol{\Omega} = [\Omega_1, \ldots, \Omega_i, \ldots, \Omega_n]^{\mathrm T}\), where \(\Omega_i \in \{0, 1\}\), i.e., element  \(x_i\) is selected if \(\Omega_i=1\), while it is discarded if \(\Omega_i=0\). For the point cloud downsampling problem, given an input point cloud \( \mathbf{P}_{in} = \{ p_i \}_{i=1}^n \), a contribution score for each point, and an integer \(k\) (\(k<n\)), our goal is to find the index vector \(\boldsymbol{\Omega}\) such that 

\begin{equation}
\label{EQUa1}
\footnotesize
\Omega_i = \begin{cases}
1, & \text{if the contribution score of } p_i \text{ is among the Top-\textit{k} in } \mathbf{P}_{in} \\
0, & \text{otherwise.}
\end{cases}
\end{equation}

\noindent Specifically, we first extract the features of each point in \(\mathbf{P}_{in}\) through the feature embedding (FE) module; then we emphasize the attractive features and suppress less important ones globally through the cascade attention (CA) module, and further calculate the contribution score of each point through the contribution scoring (CS) module, and then we retain the Top-\(k\) points with the highest contribution through the Top-\(k\) operator to achieve downsampling.

The computational process of the Top-\(k\) operator can be described as the solution process of the optimal transmission problem \cite{ref21}. Given two discrete distributions defined on supports \(\boldsymbol{\mathcal{A}} = \{a_i\}_{i=1}^n\) and \(\boldsymbol{\mathcal{B}} = \{b_j\}_{j=1}^s\), respectively. For a typical Top-\(k\) operator, \(s=2\), i.e., \(b_1=1\) indicates an element of \(\boldsymbol{\mathcal{A}}\) belongs to Top-\(k\), while \(b_2=0\) indicates an element of \(\boldsymbol{\mathcal{A}}\) does not belong to Top-\(k\). Denote \(\mathbb{P}(\{a_i\}) = \mu_i\) and \(\mathbb{P}(\{b_j\}) = \upsilon_j\), and let \(\boldsymbol{\mu} = [\mu_1, \mu_2, \ldots, \mu_n]^{\mathrm{T}}\) and \(\boldsymbol{\upsilon} = [\upsilon_1, \ldots, \upsilon_s]^{\mathrm{T}}\). The Top-\(k\) operator can be obtained by solving an optimal transmission problem \cite{ref22} which seeks to find the least costly transmission scheme between two probability distributions:

\begin{equation}
\label{EQUa2}
\mathbf{\Gamma}^* = \arg \min_{\mathbf{\Gamma} \geq 0} \langle \mathbf{C}, \mathbf{\Gamma} \rangle, \quad \text{s.t.,} \quad \mathbf{\Gamma} \mathbf{1}_s = \boldsymbol{\mu}, \mathbf{\Gamma}^{\mathrm{T}} \mathbf{1}_n = \boldsymbol{\upsilon},
\end{equation}

\noindent where \(\langle \mathbf{C}, \mathbf{\Gamma} \rangle = \sum_{i=1}^n \sum_{j=1}^s C_{i,j} \Gamma_{i,j}\), \(\mathbf{\Gamma}\) denotes a transmission matrix, \(\mathbf{C}\) denotes a cost matrix, and \({\mathbf{1}_s}\) and \({\mathbf{1}_n}\) denote column vectors consisting of \(s\) and \(n\) ones, respectively.

To parameterize the Top-\(k\) operator using \(\mathbf{\Gamma}^*\), we set \(\boldsymbol{\mathcal{A}}=\mathbf{P}_{in}\) and \(\boldsymbol{\mathcal{B}}=\{1,0\}\), with \(\boldsymbol{\mu}\), \(\boldsymbol{\upsilon}\) defined as

\begin{equation}
\label{EQUa3}
\boldsymbol{\mu} = \frac{\mathbf{1}_n}{n}, \boldsymbol{\upsilon} = \left[\frac{k}{n}, \frac{n-k}{n}\right],
\end{equation}

\noindent That is, the relationship between the output \(\boldsymbol{\Omega}\) of the Top-\(k\) operator and \(\mathbf{\Gamma}^*\) for \(\mathbf{P}_{in}\) is \(\mathbf{\Omega} = n \mathbf{\Gamma}^* \cdot [1, 0]^{\mathrm{T}}\) \cite{ref21}, where “\(\cdot\)” denotes the dot product. Since the Top-\(k\) operator is not differentiable, end-to-end network training is not possible. To overcome this problem, entropy regularization can be used \cite{ref21}

\begin{equation}
\label{EQUa4}
\footnotesize
{\mathbf{\Gamma}^{*,\varepsilon} = \arg \min_{\mathbf{\Gamma} \geq 0} \langle \mathbf{C}, \mathbf{\Gamma} \rangle + \varepsilon H(\mathbf{\Gamma}), \quad \text{s.t.,} \quad \mathbf{\Gamma} \mathbf{1}_s = \boldsymbol{\mu}, \mathbf{\Gamma}^{\mathrm{T}} \mathbf{1}_n = \boldsymbol{\upsilon},}
\end{equation}

\noindent where \(H(\mathbf{\Gamma}) = \sum_{i=1}^n \sum_{j=1}^k {\Gamma}_{i,j} \log {\Gamma}_{i,j}\) is the entropy regularizer, and \(\varepsilon\) is the regularization parameter. We define \(\mathbf{\Omega}^\varepsilon = n \mathbf{\Gamma}^{*,\varepsilon} \cdot [1, 0]^{\mathrm{T}}\) as a smoothed counterpart of \(\mathbf{\Omega}\) in the standard Top-\(k\) operator. Accordingly, the optimal transport-based differentiable Top-\(k\) operator is defined as the mapping from \(\mathbf{P}_{in}\) to \(\mathbf{\Omega}^\varepsilon\).

Therefore, the point cloud downsampling problem can be expressed as

\begin{equation}
\label{EQUa5}
\mathbf{P}_{sp} = \Psi(\mathbf{P}_{in} \mid \mathbf{\Theta}_{FE}, \mathbf{\Theta}_{CA}, \mathbf{\Theta}_{CS}, \mathbf{\Omega}^{\varepsilon}),
\end{equation}

\noindent where \(\Psi(\cdot)\) is the proposed CS-Net, and \(\mathbf{\Theta}_{FE}\), \(\mathbf{\Theta}_{CA}\), and \(\mathbf{\Theta}_{CS}\) denote the learnable parameters of the FE module, CA module, and CS module, respectively.

\section{Proposed Method}
For a given input point cloud \(\mathbf{P}_{in}\) with \(n\) points and a specific downstream task network, our goal is to design a learning-based sampling network that outputs a point set \(\mathbf{P}_{sp}\) with \(k\) \((k<n)\) points. The ratio \(n/k\) is called sampling ratio. Fig. 1 shows the architecture of CS-Net. We first use an FE module to capture both local and global characteristics. Then, we use a CA module to emphasize the attractive features and suppress less important ones. A CS module is then connected after the CA module to map the features into point-wise scores. These scores represent the importance of each point with respect to the downstream task and the loss function. We sort these scores and select the points with the \(k\) highest scores in the sampled point cloud \(\mathbf{P}_{sp}\) by introducing the Top-\(k\) operation. Finally, \(\mathbf{P}_{sp}\) is fed into the task network for the specific downstream task. In the above procedure, \(k\) can be set arbitrarily, i.e., the proposed network can adapt to various sampling ratios.
\begin{figure}
\centering
\includegraphics[width=3.2in]{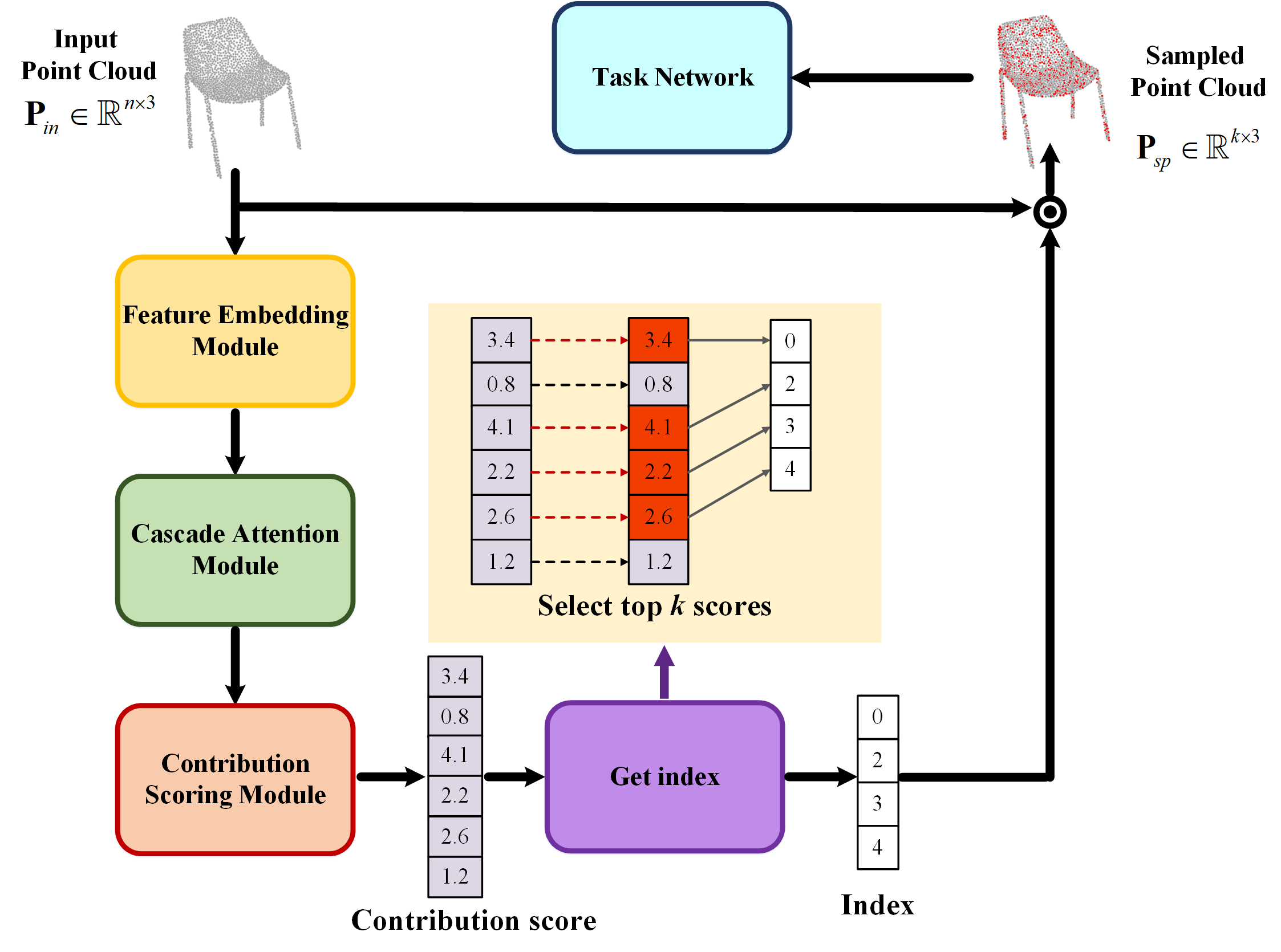}
\vspace{-10pt}
\caption{Architecture of CS-Net.We first use an FE module to capture both local and global characteristics. Then, we use a CA module to emphasize the attractive features and suppress less important ones. A CS module is then connected after the CA module to map the features into point-wise scores.These scores represent the importance of each point with respect to the downstream task and the loss function. We sort these scores and select the points with the \(k\) highest scores in the sampled point cloud \(\mathbf{P}_{sp}\) by introducing the Top-\(k\) operation.}
\label{FIG1}
\vspace{-10pt}
\end{figure}
\begin{figure}
\centering
\includegraphics[width=3.3in]{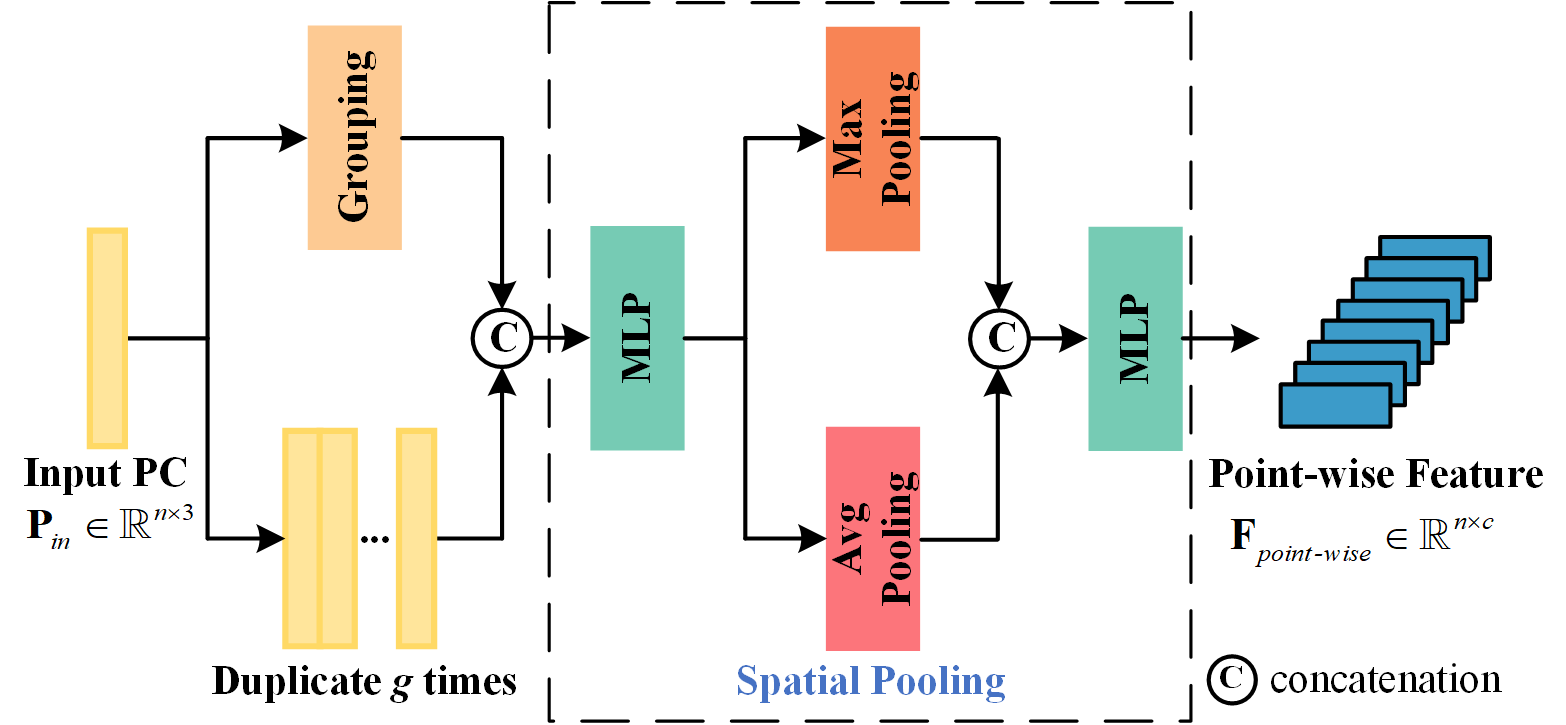}
\vspace{-8pt}
\caption{Proposed FE module. The features of each point in \(\mathbf{P}_{in}\) are extracted through the FE module.}
\label{FIG2}
\vspace{-18pt}
\end{figure}

\subsection{Feature Embedding Module}
The goal of the feature embedding module (Fig. 2) is to obtain a point-wise feature map for an input point cloud. First, we use a grouping layer \cite{ref5} to extract local features from the input point cloud based on a grouping operation. The grouping operation is defined as follows. For a point \(\bm{p}_i\) \((i=1, …, n)\) in \(\mathbf{P}_{in}\), we find its \(g\) nearest neighbors \(\bm{p}_{i,j}\), \((j=1,…,g)\) in \(\mathbf{P}_{in}\). We then compute the difference between \(\bm{p}_i\) and each of its neighbors, \(\bm{p}_{i,j}\). The relative coordinates (\(\bm{p}_{i,j}-\bm{p}_i\)) are used in the grouping operation.

The grouping layer \(\mathbf{F}_{group}\) for the input point cloud is obtained by applying the grouping operation to each point. Then, we replicate the input point cloud \(g\) times and combine the result with  \(\mathbf{F}_{group}\) to effectively preserve global features. A multi-layer perceptron (MLP) is then used to map the feature into a hyper-space to obtain  \(\mathbf{F}_{combine}\).

Next, a pooling operation is introduced to reduce the computation cost and make the network permutation-equivariant. Max pooling excels at preserving local details. On the other hand, average pooling effectively aggregates features across different channels but may lose local details \cite{refcb}. Therefore, we introduce a spatial pooling layer that combines the benefits of max pooling and average pooling:

\begin{equation}
\label{EQU1}
\mathbf{F}_{point-wise}=\xi ({ {\max ( \mathbf{F}_{combine} ),{\rm{avg}}(\mathbf{F}_{combine})}}),
\end{equation}

\noindent where max\((\cdot)\) represents the max pooling layer, avg\((\cdot)\) represents the average pooling layer, and \(\xi ( \cdot )\) represents an MLP. 
\begin{figure}
\centering
\includegraphics[width=3.2in]{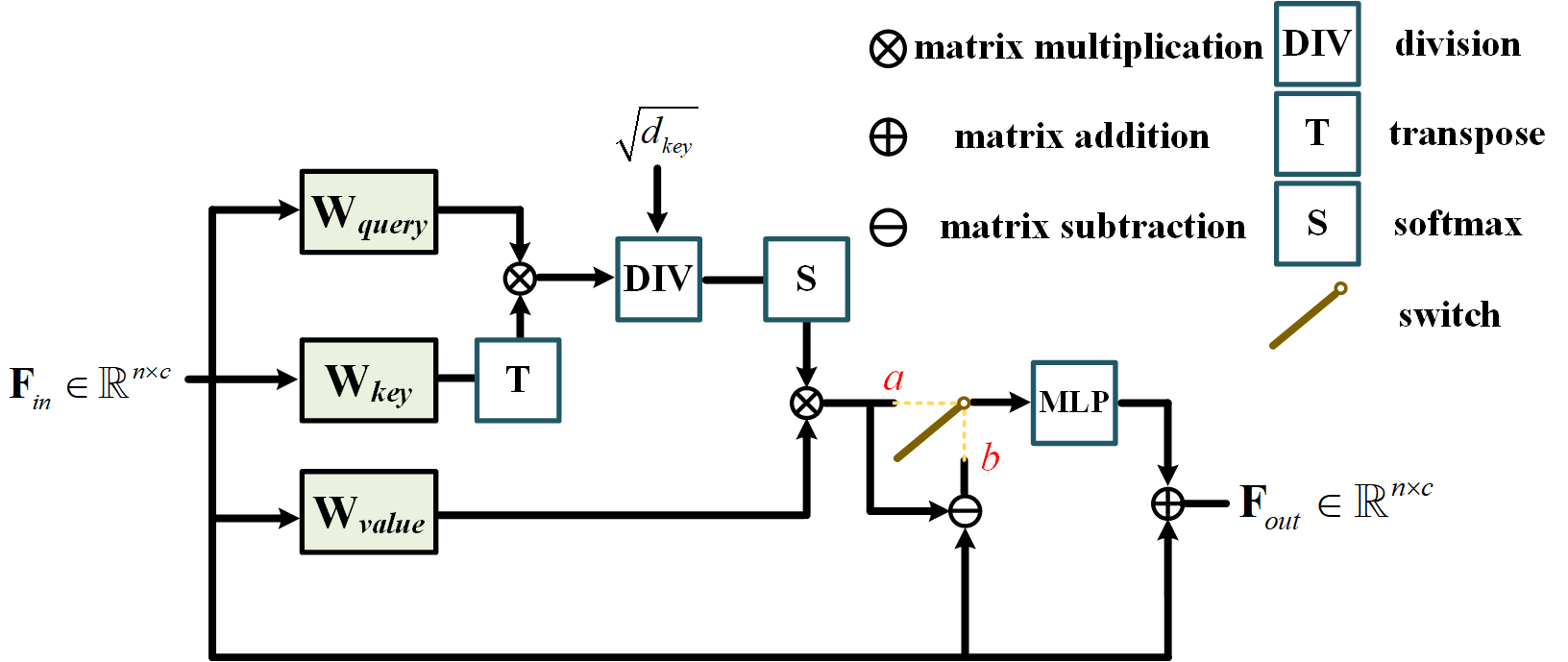}
\vspace{-8pt}
\caption{Self-attention (switch connected at side a) and offset-attention (switch connected at side b).}
\label{FIG3}
\vspace{-15pt}
\end{figure}
\vspace{-25pt}
\subsection{Cascade Attention Module}
As we aim at selecting points from the input point cloud, we use a self-attention (SA) mechanism \cite{ref23} that identifies and captures the most relevant and informative points during training. Our self-attention mechanism is given by 

\begin{align}
\label{EQU23}
(\mathbf{Q},\mathbf{K},\mathbf{V}) = {\mathbf{F}_{in}} \cdot ({\mathbf{W}_{query}},{\mathbf{W}_{key}},{\mathbf{W}_{value}}), \\
{\mathbf{F}_{sa}} = softmax\left( {\frac{{\mathbf{Q} \cdot \mathbf{K^T}}}{{\sqrt {{d_{key}}} }}} \right) \cdot \mathbf{V},
\end{align}

\noindent where \(\mathbf{F}_{in}\) denotes the input feature, \({\mathbf{W}_{query}}\), \({\mathbf{W}_{key}}\), and \({\mathbf{W}_{value}}\) are learnable \(c \times c\) matrices, and \(d_{key} = c\). The self-attention layer can be written as

\begin{equation}
\label{EQU4}
\mathbf{F}_{out}=\gamma ({\mathbf{F}_{sa}})+{\mathbf{F}_{in}},
\end{equation}

\noindent where \(\gamma (\cdot)\) denotes an MLP. However, the self-attention layer falls short in addressing the challenge of information loss as the network depth increases\cite{ref24}. By taking the difference between the attention features and the input features into account, as shown in Fig. 3, we use offset-attention (OA) \cite{ref25} to modify the feature as follows

\begin{equation}
\label{EQU5}
\mathbf{F}_{out}=OA({\mathbf{F}_{in}})=\gamma ({\mathbf{F}_{in}}-{\mathbf{F}_{sa}})+{\mathbf{F}_{in}}.
\end{equation}

As neural networks grow deeper, they may not be able to preserve all the relevant information. To address this issue, we propose CA to combine information from earlier layers with that from subsequent layers. Specifically, CA consists of three skip connected OA layers followed by a concatenation along the feature dimension (Fig. 4):

\begin{align}
\label{EQU6789}
&{\mathbf{F}_{oa_1}}=OA({\mathbf{F}_{point-wise}})\\
&{\mathbf{F}_{oa_2}}=OA({\mathbf{F}_{oa_1}})\\
&{\mathbf{F}_{oa_3}}=OA({\mathbf{F}_{oa_2}})\\
&{\mathbf{F}_{concat}}=concat({\mathbf{F}_{oa_1}}, {\mathbf{F}_{oa_2}}, {\mathbf{F}_{oa_3}}),
\end{align}

\noindent where \(\mathbf{F}_{oa_i}\) \((i = 1,2,3)\) denotes the output feature of the \(i\)-th OA layer, and \(\mathbf{F}_{concat}\) is the concatenated feature.
\begin{figure*}
\centering
\includegraphics[width=6.6in]{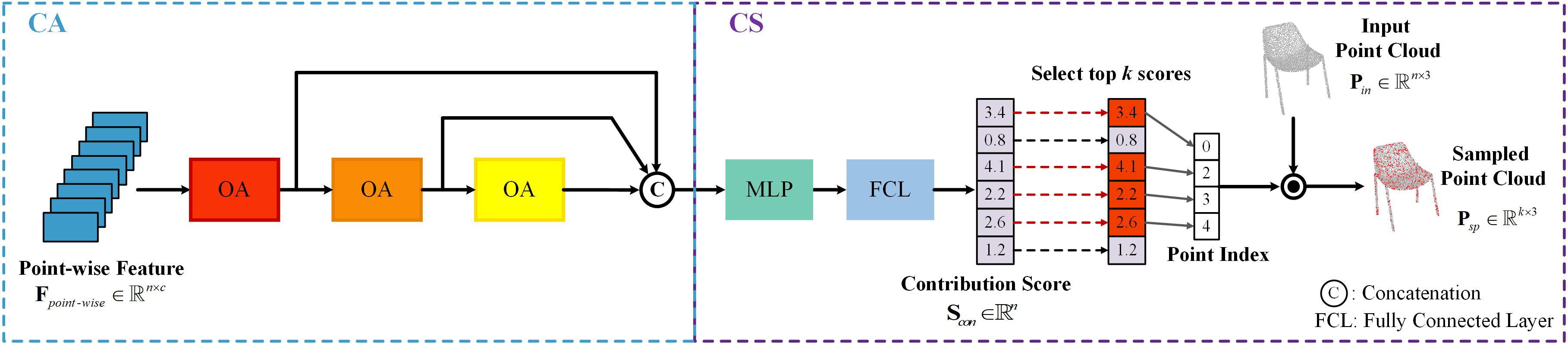}
\vspace{-6pt}
\caption{Architecture of the CA module and CS module. The CA module is used to emphasize the attractive features and suppress less important ones. The CS module is then connected after the CA module to map the features into point-wise scores. By introducing the Top-\(k\) operation, the points with the \(k\) highest scores are sorted and selected.}
\label{FIG4}
\vspace{-16pt}
\end{figure*}
\vspace{-15pt}

\subsection{Contribution Scoring Module}
Existing selective methods mostly rely on a selective matrix for sampling, which may lead to a differentiability problem and repetitive selection for certain points. To address these challenges, we convert the problem of selecting the largest \textit{k} elements into an optimal transport-based differentiable Top-\(k\) operator, as analysed in Section III. The goal of the CS module (Fig. 4) is to sort and select the points according to their contribution scores, thus ensuring that there are no duplicate points in the sampled point cloud. The concatenated feature \(\mathbf{F}_{concat}\) is mapped to the point-wise contribution scores \(\mathbf{S}_{con}\) which is used to evaluate the contribution of each point. The mapping is achieved via an MLP followed by three fully connected layers:

\begin{equation}
\label{EQU13}
{\mathbf{S}_{con}}=FC(\rho({\mathbf{F}_{concat}})),
\end{equation}

\noindent where \(\rho (\cdot)\) represents the MLP, and \(FC(\cdot)\) represents the fully connected layers. Each element in \({\mathbf{S}_{con}}\) quantifies the importance to the downstream task. The larger the score, the more important the corresponding feature is. We then select the \textit{k} elements in \({\mathbf{S}_{con}}\) with the highest scores and get the corresponding point indices. Finally, the sampled point cloud \({\mathbf{P}_{sp}}\) is obtained by referencing the indices of the selected downsampled points from the input point cloud \({\mathbf{P}_{in}}\). 
\vspace{-15pt}
\subsection{Loss Function}
In the proposed network, we use the joint loss

\begin{equation}
\label{EQU14}
{L}_{total}=\alpha{L}_{emd}({\mathbf{P}_{sp}},{\mathbf{P}_{in}})+\beta{L}_{task}({\mathbf{P}_{sp}}),
\end{equation}

\noindent where \({L}_{task}(\cdot)\) is designed to encourage the network to learn a downsampled point set that is optimized for the specific downstream task, and \({L}_{emd}(\cdot)\) is the EMD loss. \({L}_{emd}(\cdot)\) aims to minimize the distance between \({\mathbf{P}_{in}}\) and \({\mathbf{P}_{sp}}\), ensuring their similarity. It is defined as follows \cite{ref32}: 

\begin{eqnarray}
\begin{aligned}
\label{EQU15}
&{L}_{emd}({\mathbf{P}_{sp}},{\mathbf{P}_{in}})=\\&\frac{1}{{\left| {{\mathbf{P}_{sp}}} \right|}}\mathop {\min }\limits_{\varphi :{\mathbf{P}_{sp}} \to {\mathbf{P}_{in}}} \sum\limits_{x \in {\mathbf{P}_{sp}}} {\left\| {x - \varphi (x)} \right\|_2^2}.
\end{aligned}
\end{eqnarray}

\noindent where \(\varphi\) is a bijection.
\vspace{-10pt}
\section{Experimental Results}
In this section, we study CS-Net’s effectiveness for classification, registration, sampling-based compression, surface reconstruction, and object detection. As datasets, we used ModelNet40 \cite{ref26}, PU147 \cite{ref29}, and KITTI \cite{refkitti}. ModelNet40 consists of 12,311 CAD models representing 40 categories of man-made objects. We followed the default train-test split approach, partitioning the dataset into 9,843 point clouds for training and 2,468 point clouds for testing. PU147 consists of 147 point clouds, covering a rich variety of objects, ranging from simple and smooth models (e.g., Icosahedron) to complex and highly detailed objects (e.g., Statue). We used 120 point clouds for training and 27 point clouds for testing. KITTI contains 7,481 large-scale outdoor point clouds captured by LiDAR sensors mounted on cars. To improve the network’s robustness, we used data augmentation techniques such as random rotation and scaling during the training phase. The primary hardware consisted of an Intel Core i7 7820X processor paired with an NVIDIA GeForce RTX 2080 TI GPU. 
\begin{table*}
    \centering
     \caption{Classification accuracy (\%) comparison with state-of-the-art sampling methods on ModelNet40}
     \label{tab:my_label1}
     \resizebox{17cm}{!}{
    \begin{tabular}{cccccccccccc}
    \toprule
     \textit{k}& \makecell{Sampling \\ratio} &\makecell{FPS\\\cite{ref5}} &\makecell{S-Net\\\cite{ref9}} &\makecell{SampleNet\\ \cite{ref10} }& \makecell{CO-NET\\ \cite{ref11}} & \makecell{DA-Net\\ \cite{ref12}} & \makecell{PST-NET\\ \cite{ref13}} &\makecell{MOPS-Net \\\cite{ref15}} &\makecell{SS-Net \\\cite{ref18}}&\makecell{ CAS-Net\\ \cite{ref33}} &\makecell{CS-Net}\\\hline
        512&2&88.30&87.80&88.16	&88.70&89.01	&87.94	&86.75	&87.84&89.62&\textbf{89.67}\\
        256&4&83.64&82.38&84.27&87.50& 86.24&	83.15&86.10&87.47&88.86&\textbf{89.24}\\
        128&8&70.34	&77.53	&80.75	&87.20	&85.67	&80.11	&86.05	&86.39	&87.92	&\textbf{88.48}\\\bottomrule
    \end{tabular}}  
    \vspace{-15pt}
\end{table*}

\begin{table}
    \centering
    \caption{Classification accuracy (\%) with different loss functions on ModelNet40}
     \label{tab:my_label2}
    \begin{tabular}{cccc}
    \hline
        Sampling Ratio & Cross entropy & EMD & Cross entropy + EMD \\ \hline
        2 & 87.76 & 6.19 & 89.67 \\ \hline
    \end{tabular}
\end{table}
\vspace{-20pt}
\subsection{Sampling for Classification}
For classification, we used PointNet \cite{ref27} as a task network and ModelNet40 as a dataset. Each input point cloud consisted of 1024 points. The loss function, \({L}_{task}(\cdot)\), for the classification task was the cross-entropy between the predicted labels and the ground truth labels. We trained CS-Net and PointNet jointly. During the training phase, we used a batch size of 8, conducted training over 200 epochs, and set the learning rate to 0.001. Moreover, we set \(\alpha = 1\), \(\beta=1\) in the loss function, \(c = 64\), and \(g = 32\).

We compared our method with eight state-of-the-art learning-based sampling methods: S-Net \cite{ref9}, SampleNet \cite{ref10}, CO-NET \cite{ref11}, DA-Net \cite{ref12}, PST-NET \cite{ref13}, MOPS-Net \cite{ref15}, SS-Net \cite{ref18}, and CAS-Net \cite{ref33}, which were all jointly trained with PointNet. We also included FPS \cite{ref5} as a benchmark. To adapt SS-Net to our evaluation, we modified its network structure by eliminating its fully connected layers and directing the output point cloud of the straight sampling (SS) module to PointNet for classification.

Table I presents the classification accuracy at various sampling ratios. The results show that CS-Net outperformed the other methods consistently across all sampling ratios. Note that the advantages of CS-Net became more pronounced as the sampling ratio increased. This phenomenon can be attributed to the intensifying loss of information in the downsampled point cloud as the sampling ratio increases. In such scenarios, CS-Net's ability to retain points with crucial features allowed it to maintain superior classification performance.
\begin{figure}
\centering
\includegraphics[width=3in]{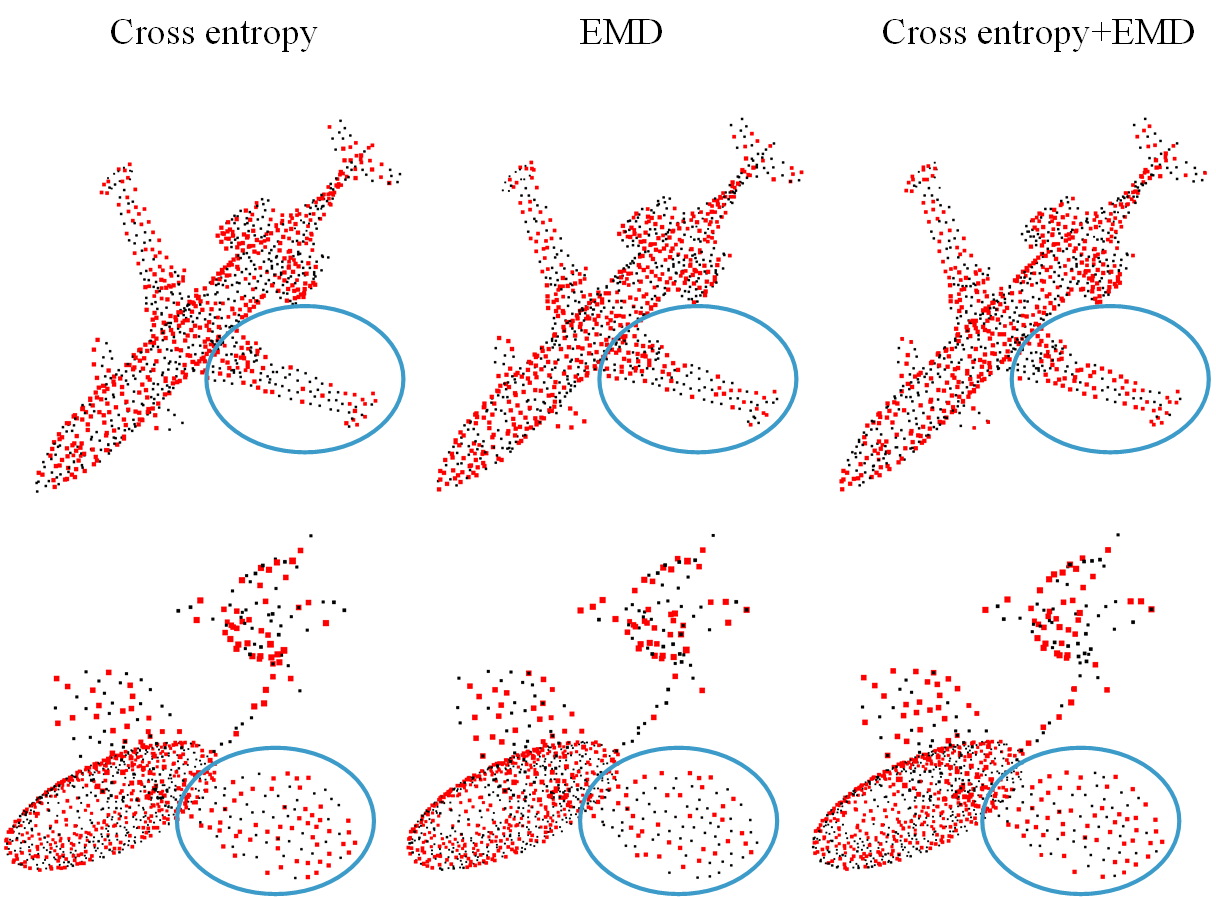}
\caption{Subjective visual comparison of sampled point clouds with different loss functions. Black points represent the original point cloud, and red points represent the sampled point cloud.}
\label{FIG5}
\end{figure}

Table II shows the effect of different loss functions on classification accuracy. Using only the cross-entropy loss resulted in high classification accuracy (87.76\%). In contrast, the accuracy obtained with only the EMD loss function was significantly lower (6.19\%). By combining cross-entropy and EMD loss, CS-Net achieved a classification accuracy of 89.67\%. This suggests that the joint loss function effectively preserved important features and helped the network adapt to the task. Fig. 5 shows the sampled point clouds obtained with different loss functions.The joint loss function best preserved both overall shape and local details of the point clouds, such as the wing of the \textit{airplane} point cloud and the leaves of the \textit{flower} point cloud.

\subsection{Sampling for Registration}
In this experiment, our approach involved two main steps. First, we used CS-Net for point cloud sampling, and then we applied the Go-ICP algorithm \cite{ref30} to align the rotated point cloud with the sampled one. To evaluate the network's performance, we considered the rotation error between the registered point cloud and the sampled point cloud. We used a batch size of 4, 300 epochs and a learning rate of 0.001 during training. We set \(\alpha = 1\), \(\beta=0\) in the loss function, \(c = 64\), and \(g = 32\) in the training phase.

Fig. 6 compares CS-Net to FPS, S-Net, and SampleNet on PU147 and ModelNet40. The input point clouds consisted of 2048 obtained by uniformly sampling the mesh surface and normalizing the unit sphere. Similar to CS-Net, S-Net and SampleNet were trained independently of the registration task. The results show that CS-Net consistently outperformed the other methods across all tested sampling ratios. This favorable result can be attributed to its ability to strategically select the most representative and significant points, which assists the Go-ICP algorithm in finding a suitable alignment solution.

\begin{table*}
    \centering
    \caption{Bjøntegaard-Delta peak signal-to-noise ratio (BD-PSNR) and Bjøntegaard-Delta  bitrate (BD-BR) performance using G-PCC as the baseline on PU147 and ModelNet40}
     \label{tab:my_label3}  
     \resizebox{16cm}{!}{
    \begin{tabular}{c|cc|cc|cc}
    \toprule
    \multirow{2}{*}{Point cloud}& \multicolumn{2}{c|}{SampleNet + G-PCC + PU-Refiner}& \multicolumn{2}{c|}{FPS + G-PCC + PU-Refiner}& \multicolumn{2}{c}{CS-Net + G-PCC + PU-Refiner}\\
    \cline{2-7}
     &BD-PSNR (dB)& BD-BR (\%) &BD-PSNR (dB) & BD-BR (\%)&BD-PSNR (dB) & BD-BR (\%)\\\hline
        11509\_Panda\_v4 & -2.50 & 199.12 & 0.77 & -12.73 & 1.19 & -16.17  \\ 
        13770\_Tiger\_v1 & -5.79 & 1232.91 & 1.08 & -14.70 & 1.41 & -19.82  \\ 
        Gramme\_aligned & -6.64 & 1075.17 & -1.03 & 7.80 & 0.14 & -1.16  \\ 
        camel & -4.65 & 560.86 & 0.90 & -5.88 & 0.84 & -10.33  \\ 
        cow & -5.33 & 1342.20 & 0.08 & -6.59 & 0.64 & -13.92  \\ 
        duck & -3.26 & 149.38 & 1.46 & -17.31 & 2.19 & -19.54  \\ 
        eight & -8.89 & 1321.28 & 0.40 & -12.57 & 1.14 & -16.93  \\ 
        elk & -2.31 & 178.72 & -0.11 & -5.29 & 0.47 & -9.26  \\ 
        fandisk & -2.71 & 192.00 & 0.81 & -3.32 & 1.14 & -5.81  \\ 
        genus3 & -6.28 & 386.05 & 1.58 & -11.23 & 1.81 & -10.57  \\ 
        horse & -7.79 & 451.22 & 1.21 & -10.02 & 1.04 & -12.11  \\ 
        kitten & -4.71 & 150.46 & 1.51 & -13.18 & 1.66 & -14.41  \\ 
        m32 & -4.11 & 284.57 & 0.51 & -13.98 & 0.36 & -13.55  \\ 
        m329 & -6.84 & 801.66 & 0.08 & -2.81 & 0.69 & -6.81  \\ 
        m333 & -7.62 & 419.95 & 1.64 & -11.06 & 1.57 & -13.61  \\ 
        m355 & -3.44 & 175.59 & 1.56 & -9.37 & 1.71 & -11.06  \\ 
        m60 & -8.01 & 105.89 & 0.70 & -18.48 & 1.10 & -17.58  \\ 
        m88 & -7.02 & 99.8 & 0.30 & -4.89 & 1.07 & -11.53  \\ 
        star & 0.41 & -15.51 & 1.60 & -18.18 & 2.28 & -22.78  \\ 
        pig & -8.70 & 528.25 & 0.59 & -9.04 & 1.16 & -15.73  \\ \hline
        PU147 average	&-5.31&481.98&0.78&-9.64&\textbf{1.18}&\textbf{-13.14}\\\hline
        ariplane & -3.41 & 123.24 & 0.07 & -2.12 & 0.89 & -8.06 \\ 
        bathtub & -2.56 & 242.92 & 0.38 & -4.97 & 0.87 & -18.29 \\ 
        bed & -7.63 & 1249.62 & -0.22 & 4.93 & 1.27 & -18.14 \\ 
        bench & -2.46 & 350.81 & 0.76 & -9.68 & 1.09 & -3.01 \\ 
        bookshelf & -4.31 & 742.28 & 0.15 & -9.2 & 0.02 & -1.92 \\ 
        bowl & -5.35 & 1231.55 & 0.41 & -3.22 & 0.81 & -20.13 \\  
        cup & -5.37 & 1332.61 & 0.03 & -3.97 & 0.42 & -14.73 \\ 
        curtain & -7.41 & 1516.55 & -0.06 & 1.98 & 0.47 & -27.99 \\ 
        door & -6.74 & 1311.41 & -0.27 & 24.18 & 0.84 & -12.46 \\ 
        flower\_pot & -2.13 & 451.34 & 0.56 & -3.02 & 0.38 & -12.43 \\ 
        guitar & -6.37 & 1334.45 & -0.18 & 2.96 & -0.09 & 1.99 \\ 
        lamp & -2.83 & 451.39 & 0.24 & -3.81 & 1.18 & -26.55 \\ 
        mantel & -2.71 & 328.91 & 1.27 & -18.42 & 1.23 & -16.22 \\ 
        person & -5.12 & 1071.7 & 0.05 & -2.37 & -0.08 & -1.84 \\ 
        sink & -3.92 & 238.16 & 0.23 & -3.89 & 0.52 & -14.06 \\ 
        table & -3.41 & 178.53 & 0.75 & -20.18 & 0.37 & -2.98 \\ 
        vase & -3.77 & 418.24 & 0.08 & -4.36 & 0.16 & -3.83 \\ 
        xbox & -2.91 & 315.12 & 0.72 & -9.2 & 1.1 & -17.21 \\ \hline
        ModelNet40 average & -4.36 & 716.05 & 0.28 & -3.58 & \textbf{0.64 }& \textbf{-12.10} \\ \bottomrule
    \end{tabular}}
    \vspace{-15pt}
\end{table*}
\begin{figure}
\centering
\includegraphics[width=3.1in]{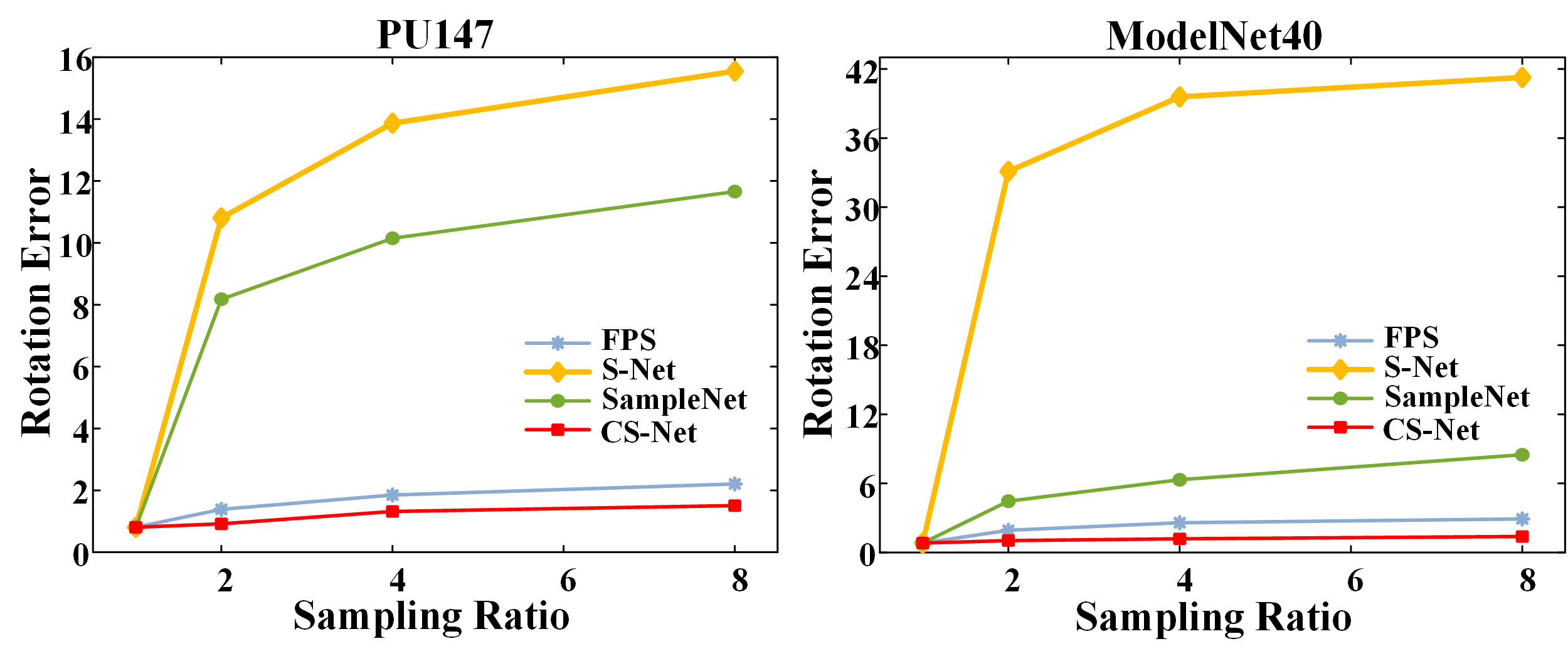}
\caption{Average rotation error on PU147 and ModelNet40.}
\label{FIG6}
\end{figure}
\begin{figure}
\centering
\includegraphics[width=3.1in]{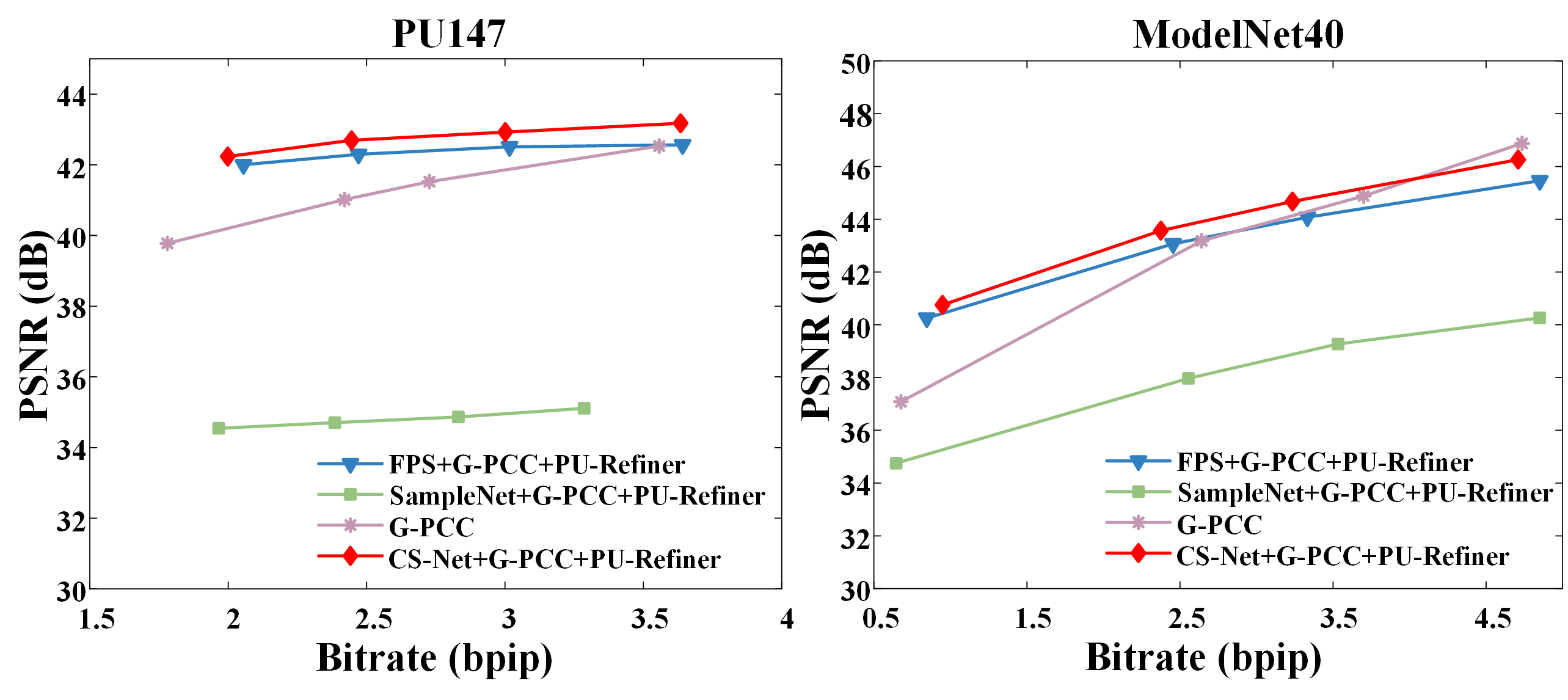}
\caption{Rate-distortion curves on PU147 and ModelNet40.The bitrate is expressed in bits per input bit (bpip). }
\label{FIG7}
\vspace{-15pt}
\end{figure}

\subsection{Sampling for Compression}
In this section, the input point clouds were obtained as in Section V-B. For this task, our approach involved a multi-step process. First, CS-Net was used to sample the input point cloud. Then, the downsampled point cloud was encoded using the Moving Picture Experts Group (MPEG) geometry-based point cloud compression (G-PCC) reference software \cite{ref28}. Finally, the reconstructed downsampled point cloud was upsampled using PU-Refiner \cite{ref3}, a state-of-the-art upsampling network. The network's efficacy was evaluated based on the distortion between the initial point cloud and its upsampled version.

Due to constraints imposed by the G-PCC architecture, CS-Net and PU-Refiner were trained independently. During training, we used a batch size of 4, spanning a total of 400 epochs, while maintaining a learning rate of 0.001. We set the parameters to \(\alpha = 1\), \(\beta=0\), \(c = 64\), and \(g = 32\) throughout the training procedure. We used a pretrained model for PU-Refiner.

For ease of processing, during the compression phase, we transformed the point cloud into a voxelized representation within a cubic space with a side length of 512 units, resulting in integer geometric coordinates. After downsampling, the resulting point cloud was encoded and decoded using the TMC13v19.0 G-PCC reference software. Note that, due to the use of lossy compression, the number of points in the decoded point cloud may vary. Finally, the decoded point cloud was upsampled by a factor of four.

Fig. 7 and Table III compare the rate-distortion performance of FPS, SampleNet, and CS-Net. The G-PCC encoder, which directly encodes the input point clouds, served as the baseline for comparison. Both CS-Net and FPS enhanced G-PCC when integrated into the proposed framework, particularly at low bitrates. Note that the compression approach using CS-Net displayed superior performance in terms of geometric PSNR improvement and computational efficiency compared to the one using FPS. This advantage can be attributed to the proficiency of the proposed downsampling mechanism in preserving sophisticated details and shape characteristics of the input point cloud. Fig. 8 highlights the aforementioned advantages of the proposed downsampling mechanism.

\vspace{-15pt}
\subsection{Sampling for Surface Reconstruction}
\vspace{-5pt}
In this section, the input point clouds were obtained as in Section V-B. For surface reconstruction, we used CS-Net for point cloud sampling and then converted the sampled point cloud into a mesh with the screened Poisson surface reconstruction method \cite{ref31}. In the training phase, we used a batch size of 4 and 300 epochs with a fixed learning rate of 0.001. We set \(\alpha = 1\), \(\beta=0\), \(c = 64\), and \(g = 32\) throughout the training process.
\begin{figure*}
\centering
\includegraphics[width=6.5in]{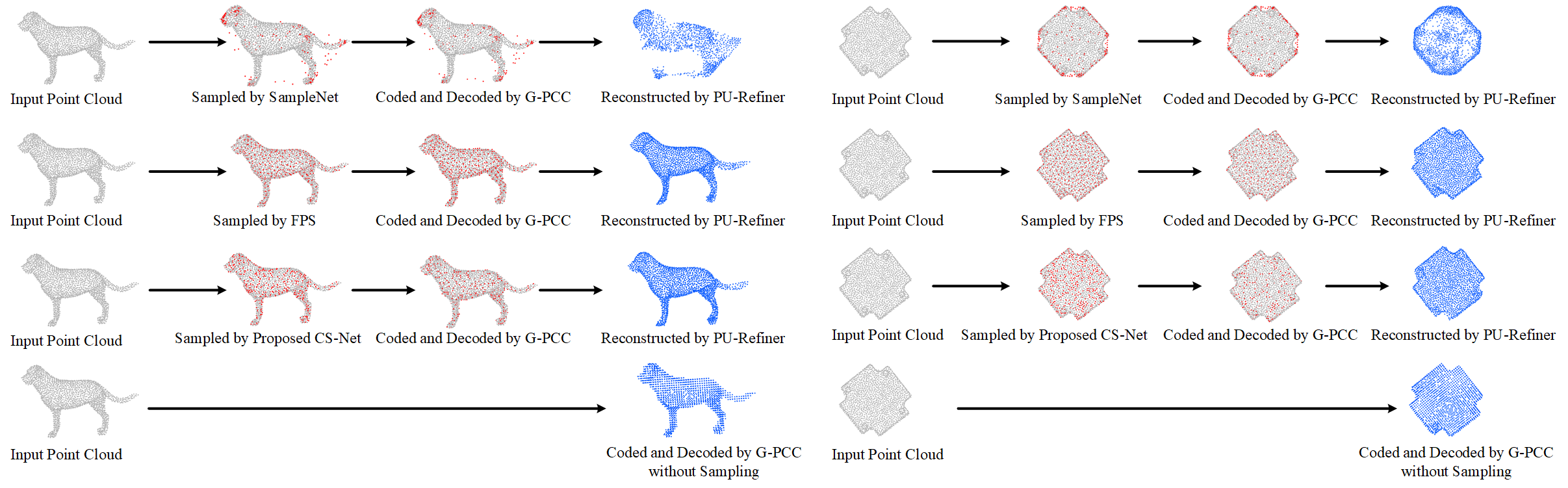}
\caption{Visualization of the sampling-based compression task.}
\label{FIG8}
\end{figure*}

\begin{figure*}
\centering
\includegraphics[width=6.5in]{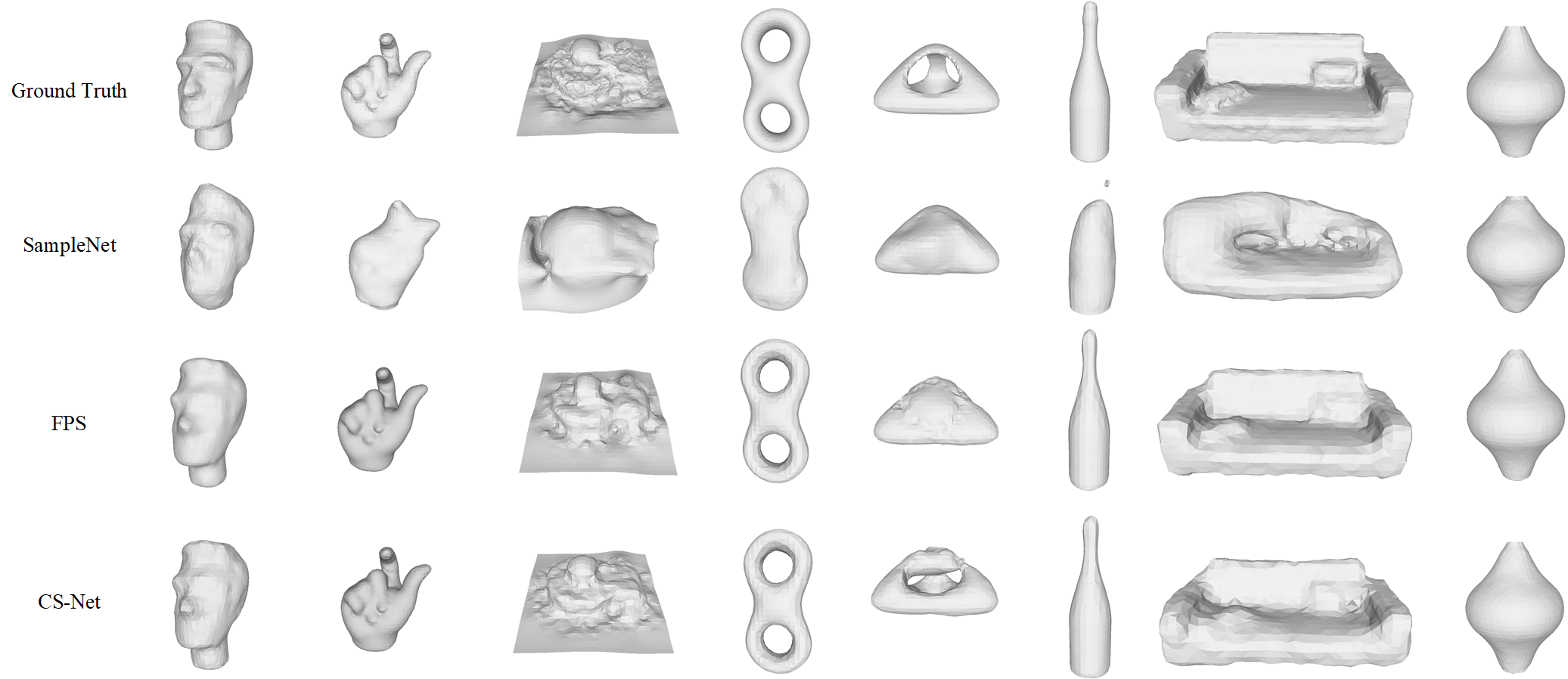}
\caption{Surface reconstruction comparison of CS-Net with SampleNet, FPS on PU147 and ModelNet40. The point clouds from left to right are \textit{head}, \textit{m333}, \textit{casting}, \textit{eight},\textit{ genus}, \textit{bottle}, \textit{sofa}, and \textit{vase}, with the first 5 from PU147 and the rest from ModelNet40.}
\label{FIG9}
\vspace{-15pt}
\end{figure*}

In Fig. 9, we compare our approach to SampleNet \cite{ref10} and FPS \cite{ref5}. SampleNet was trained independently of the downstream task. The findings show that, unlike SampleNet, both CS-Net and FPS excelled in retaining the structural characteristics of the input point cloud. Furthermore, owing to CS-Net's ability of preserving important points, the reconstructed mesh from CS-Net exhibited more sophisticated details compared to the mesh reconstructed using FPS.

\begin{figure*}
\centering
\includegraphics[width=6in]{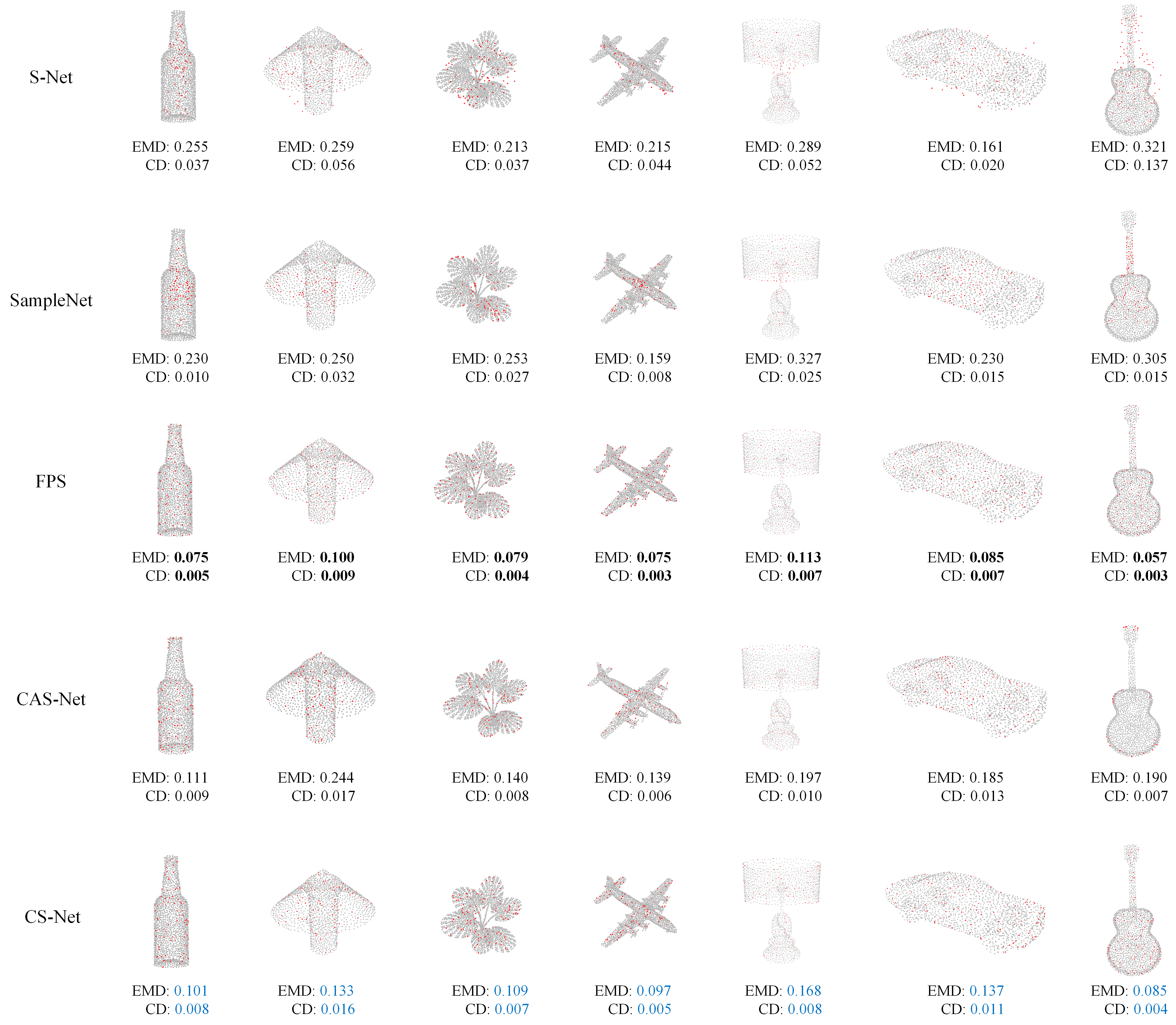}
\caption{Comparison between CS-Net and state-of-the-art sampling methods. The sampled points are marked in red.The lowest EMD and Chamfer distance (CD) between the input point cloud and the sampled point cloud are highlighted in bold. The second-lowest distances are highlighted in blue.}
\label{FIG10}
\end{figure*}
\vspace{-10pt}
\subsection{Sampling for Object Detection}
\vspace{-5pt}
To demonstrate the effectiveness of CS-Net in real-world applications where the point clouds may be very large, we assessed its performance for object detection on the KITTI dataset. We randomly selected 1,000 point clouds for training from the 7,481 point clouds in the dataset. Due to memory constraints, we split each point cloud into patches of 8,192 points (CS-Net\_w8192) and 2048 points (CS-Net\_w2048). During training, we used a batch size of 1 and 200 epochs with a fixed learning rate of 0.001. We set \(\alpha = 1\), \(\beta=0\), \(c = 64\), and \(g = 32\).
\begin{table}
    \centering
    \caption{Performance comparison on the KITTI validation set with AP calculated by 11 recall positions for car class}
    \label{tab:my_label4}  
    \resizebox{7.6cm}{!}{
    \begin{tabular}{cccc}
    \toprule
        \multirow{2}{*}{Method}& \multicolumn{3}{c}{\({\rm A}{{\rm P}_{\rm R11\_3{\rm{D}}}}\) (\%)}\\
        \cline{2-4}
        & Easy & Moderate & Hard  \\\hline 
        CS-Net\_w2048+Voxel R-CNN & 88.54 & 77.69 & 75.29  \\ 
        CS-Net\_w8192+Voxel R-CNN & 88.48 & 77.82 & 75.43  \\ 
        Voxel R-CNN & 89.41 & 84.52 & 78.93 \\ \bottomrule
    \end{tabular}}
    \vspace{-20pt}
\end{table}

As a task network, we used Voxel R-CNN \cite{refvoxel}. Following common practice, we divided the KITTI dataset into a training set (3712 point clouds) and a validation set (3769 point clouds). As the test set lacks labels, we used the validation set for testing. Table IV compares the average precision (AP) for the original point cloud and the sampled point cloud for the class “Car” with a 0.7 IoU threshold. To demonstrate the generalizability of CS-Net, we did not retrain Voxel R-CNN. The AP of “Voxel R-CNN” in Table IV was obtained by training Voxel R-CNN on the original training set and testing on the original validation set. The AP of both “CS-Net\_w8192 + Voxel R-CNN” and “CS-Net\_w2048 + Voxel R-CNN” was close to that of Voxel R-CNN, even though we did not retrain Voxel R-CNN with the downsampled point clouds.

\subsection{Shape Preservation}
Fig. 10 shows point clouds generated by FPS \cite{ref5}, S-Net \cite{ref9}, SampleNet \cite{ref10}, CAS-Net \cite{ref33} and CS-Net. All learning-based networks were jointly trained with PointNet for classification. FPS generated the most uniformly distributed sampling results with the best subjective quality. S-Net and Sample-Net tended to favor selecting some important points, which resulted in a significant loss of local details and led to larger visual distortions. Benefited from the Top-\(k\) operation, the EMD loss, and the proposed network structure, CS-Net excelled at preserving the shape of smooth point clouds while adeptly capturing sophisticated local details in complex point clouds. Fig. 10 also gives the Chamfer distance (CD) and EMD between the input point cloud and the sampled point cloud for all methods. CD measures the dissimilarity between two sets of points by calculating for each point in one set, the minimum distance to the other set. The CD between an input point cloud \(\mathbf{P}_{in}\) and the sampled point cloud \(\mathbf{P}_{sp}\) is 
\vspace{-5pt}
\begin{eqnarray}
\begin{aligned}
\label{EQU16}
{CD}(\mathbf{P}_{in},\mathbf{P}_{sp}) = &\frac{1}{{\left| {\mathbf{P}_{in}} \right|}}\sum\limits_{x \in {\mathbf{P}_{in}}} {\mathop {\min }\limits_{y \in {\mathbf{P}_{sp}}} } \left\| {x - y} \right\|_2^2 + \\&\frac{1}{{\left| {\mathbf{P}_{sp}} \right|}}\sum\limits_{y \in {\mathbf{P}_{sp}}} {\mathop {\min }\limits_{x \in {\mathbf{P}_{in}}} } \left\| {y - x} \right\|_2^2.
\end{aligned}
\end{eqnarray}

\subsection{Time Complexity}
Table V compares the time complexity of CS-Net, S-Net \cite{ref9}, SampleNet \cite{ref10}, and FPS \cite{ref5} on the whole ModelNet40 test set (2,468 point clouds, each with 1024 points, sampled with ratio 2). CS-Net exhibited the lowest time complexity, followed by S-Net, SampleNet, and FPS.

\begin{table}
    \centering
     \caption{Time complexity comparison}
     \label{tab:my_label5}  
     \resizebox{6cm}{!}{
    \begin{tabular}{cc}
    \toprule
        Model  & \makecell{Processing time (s) over \\the whole ModelNet40 test set}      \\ \hline
        S-Net \cite{ref9} & 10.95  \\ 
        SampleNet \cite{ref10} & 12.85  \\ 
        FPS \cite{ref5} & 23.60  \\ 
        CS-Net & 10.88  \\ \bottomrule
    \end{tabular}}
\end{table}
\begin{table}
    \centering
    \caption{Comparison between OA, SA, and MLP for point cloud classification}
     \label{tab:my_label6}  
     \resizebox{4.3cm}{!}{
    \begin{tabular}{cc}
    \toprule
        Model & Accuracy (\%)  \\ \hline
        CS-Net (OA) & 89.67  \\ 
        CS-Net (SA) & 89.30  \\ 
        CS-Net (MLP) & 88.12  \\ \bottomrule
    \end{tabular}}
    \vspace{-15pt}
\end{table}
Fig. 11 shows the relationship between the number of input points and the computation time for three sampling ratios. The computation time shows a near-linear increase with the number of input points, making it efficient for handling large datasets.
\begin{figure}
\centering
\includegraphics[width=2.4in]{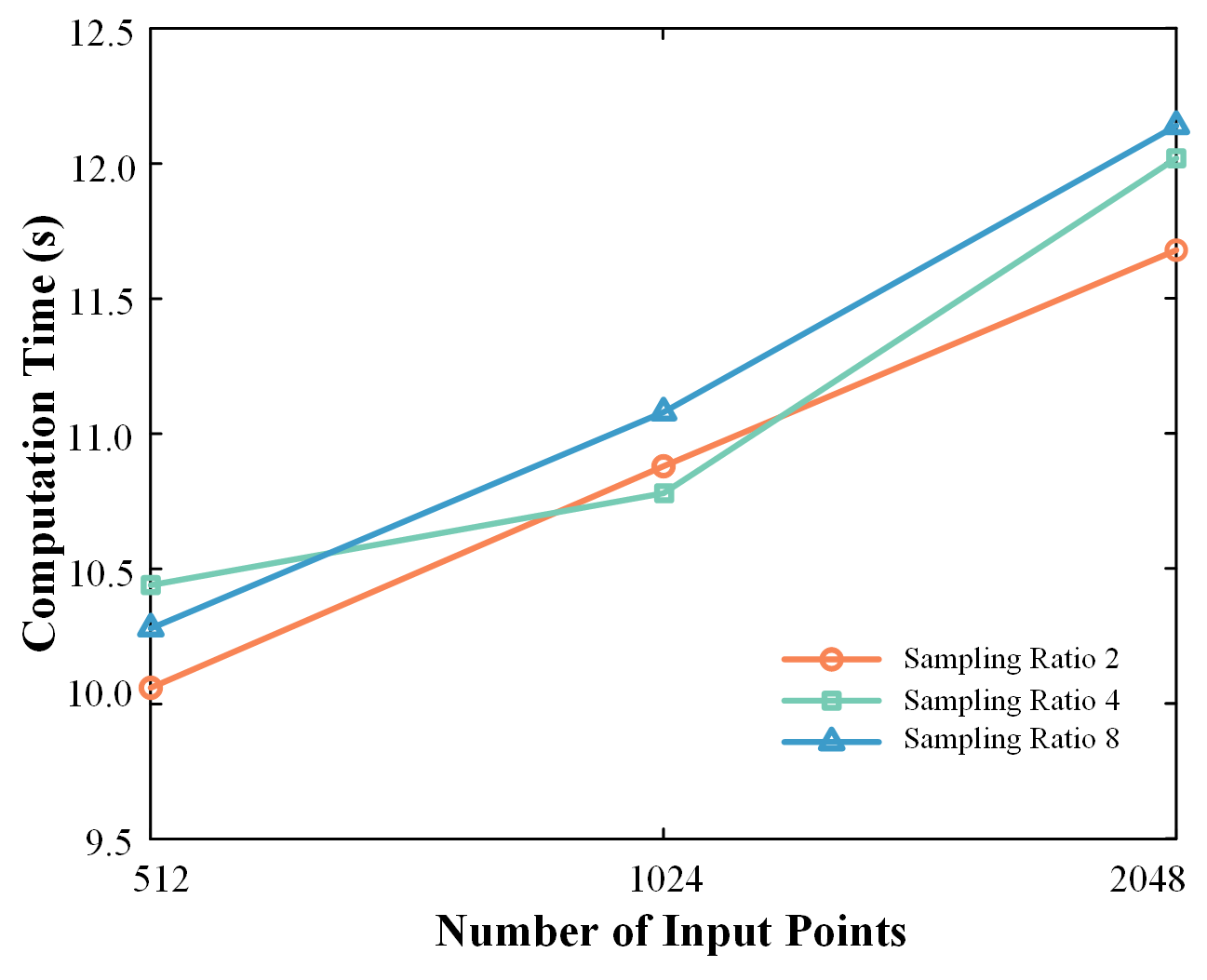}
\vspace{-10pt}
\caption{Relationship between the number of input points and the computation time for three sampling ratios. The time was computed for all 2,468 test point clouds in ModelNet40.}
\label{FIG11}
\vspace{-10pt}
\end{figure}
\begin{figure}
\centering
\includegraphics[width=3.1in]{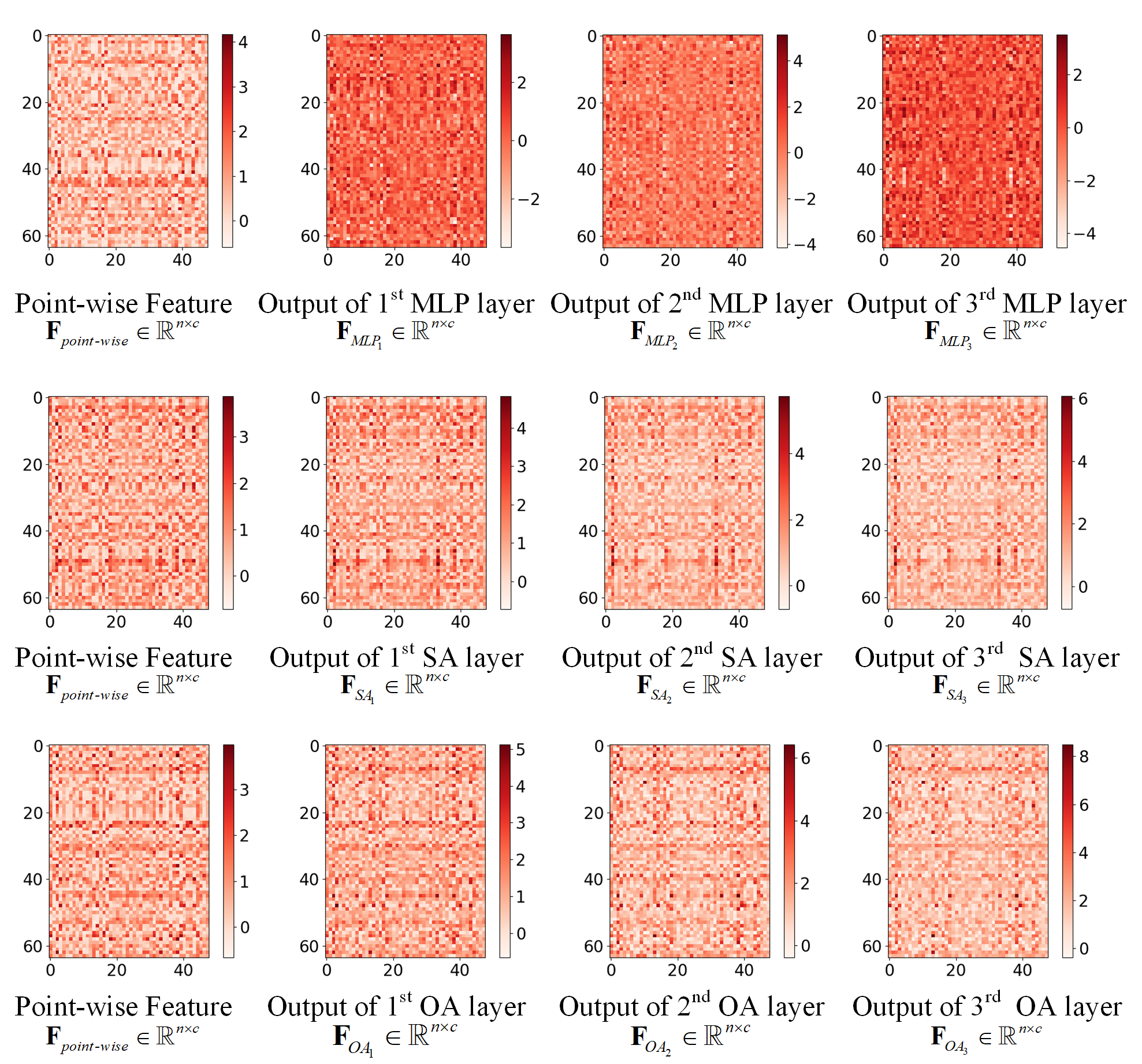}
\caption{Feature map visualization. For clarity, we randomly selected 48 points from the tested point cloud.}
\label{FIG12}
\end{figure}
\begin{figure}
\centering
\includegraphics[width=2.5in]{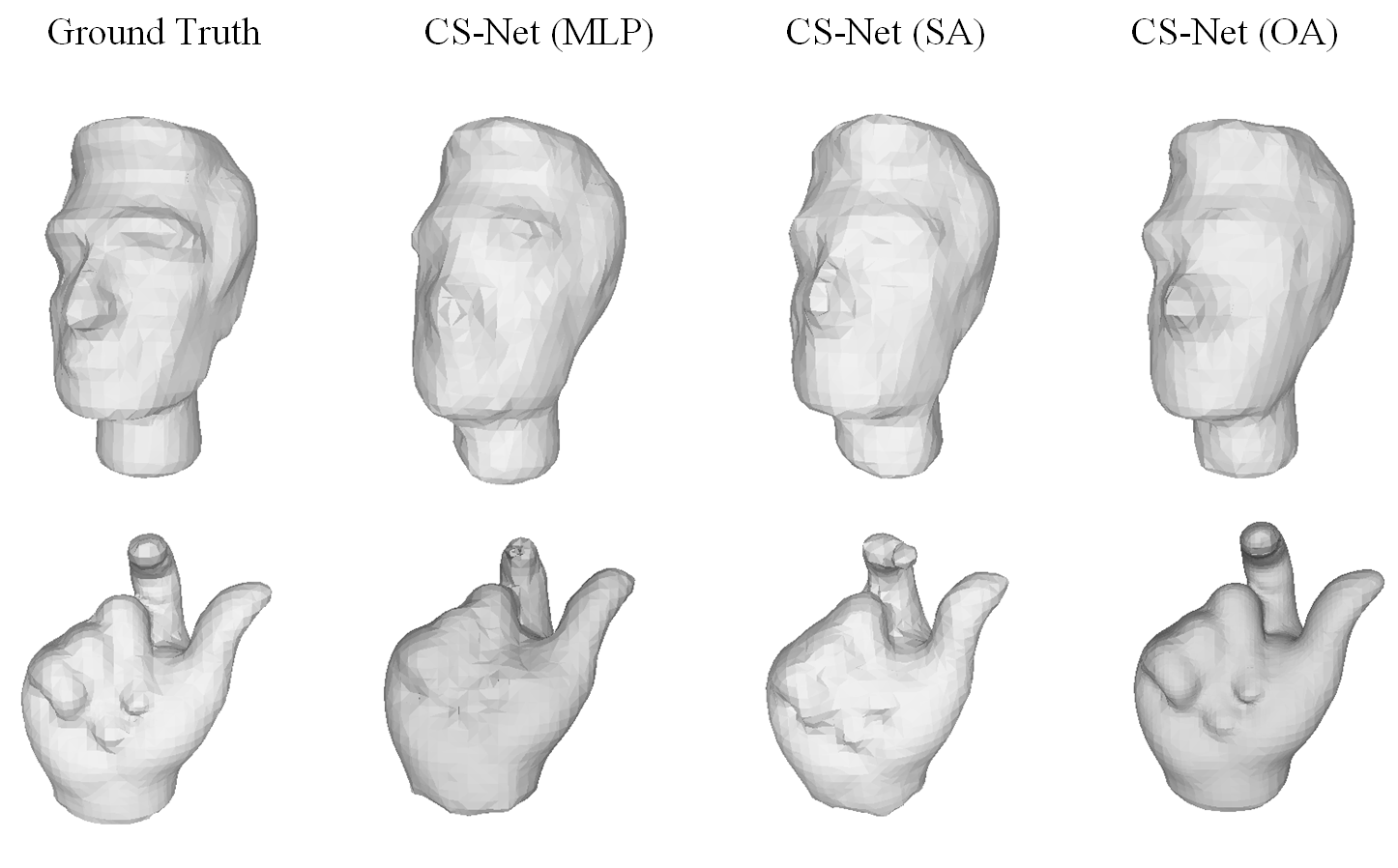}
\caption{Surface reconstruction with MLP, SA, and OA.}
\label{FIG13}
\end{figure}

\subsection{Ablation Study}
\textit{\textbf{1) Validation of OA-based Attention}}

To validate the efficacy of the OA-based attention feature extraction, we conducted two experiments: in one, we replaced the OA module with the SA module and in the other, we replaced it with a two-layer MLP module. In the two experiments, CS-Net was trained for classification with PointNet. The results in Table VI clearly demonstrate that the OA-based attention feature extraction led to more effective features, resulting in higher classification accuracy for the downsampled point clouds. The visualization of the feature maps in Fig. 12 further supports these findings. 

Notably, as the depth of the network increased, the feature map of the MLP module maintained a dense and cluttered distribution. On the other hand, for both the OA and SA modules, the features of salient points were enhanced, while those of other points were suppressed. Note that the outputs of the last two SA modules exhibit a striking similarity. This observation can be attributed to the self-attention module's constrained capacity to capture diverse feature representations and subtle variations within input features as the network depth increases.

In a similar ablation study, we compared the OA module, SA module, and MLP for surface reconstruction. The results in Fig. 13 show that surface reconstruction with OA-based attention feature extraction provided better visual results than with SA and MLP.
\begin{table}
    \centering
    \caption{Comparison of classification accuracy (\%) for different loss functions}
     \label{tab:my_label7}  
     \resizebox{7.2cm}{!}{
    \begin{tabular}{ccccc}
    \toprule
        \textit{k} & \makecell{Sampling \\ratio} & \makecell{CS-Net\\ (EMD)} &\makecell{ CS-Net \\(CD)} & \makecell{CS-Net\\ (CD+EMD)}  \\ \hline
        512 & 2 & 89.67 & 89.65 & 89.51  \\ 
        256 & 4 & 89.24 & 89.20 & 89.12  \\ 
        128 & 8 & 88.48 & 88.40 & 88.60  \\ \bottomrule
    \end{tabular}}
\end{table}

\textit{\textbf{2) Validation of EMD}}

To assess the effectiveness of the EMD loss function in terms of classification and surface reconstruction, we considered two alternative approaches. In the first one, we replaced EMD with a CD-based loss function. In the second approach, we used both CD and EMD. Table VII shows that the model using only EMD exhibited superior performance compared to the one using only CD, particularly for higher sampling ratios. Moreover, the model that incorporates both CD and EMD gave the best performance for a sampling ratio of 8 but was less effective for other ratios. This can be attributed to the divergent optimization objectives of EMD loss and CD loss, leading to fluctuations in the classification performance of the model.

\begin{figure}
\centering
\includegraphics[width=2.5in]{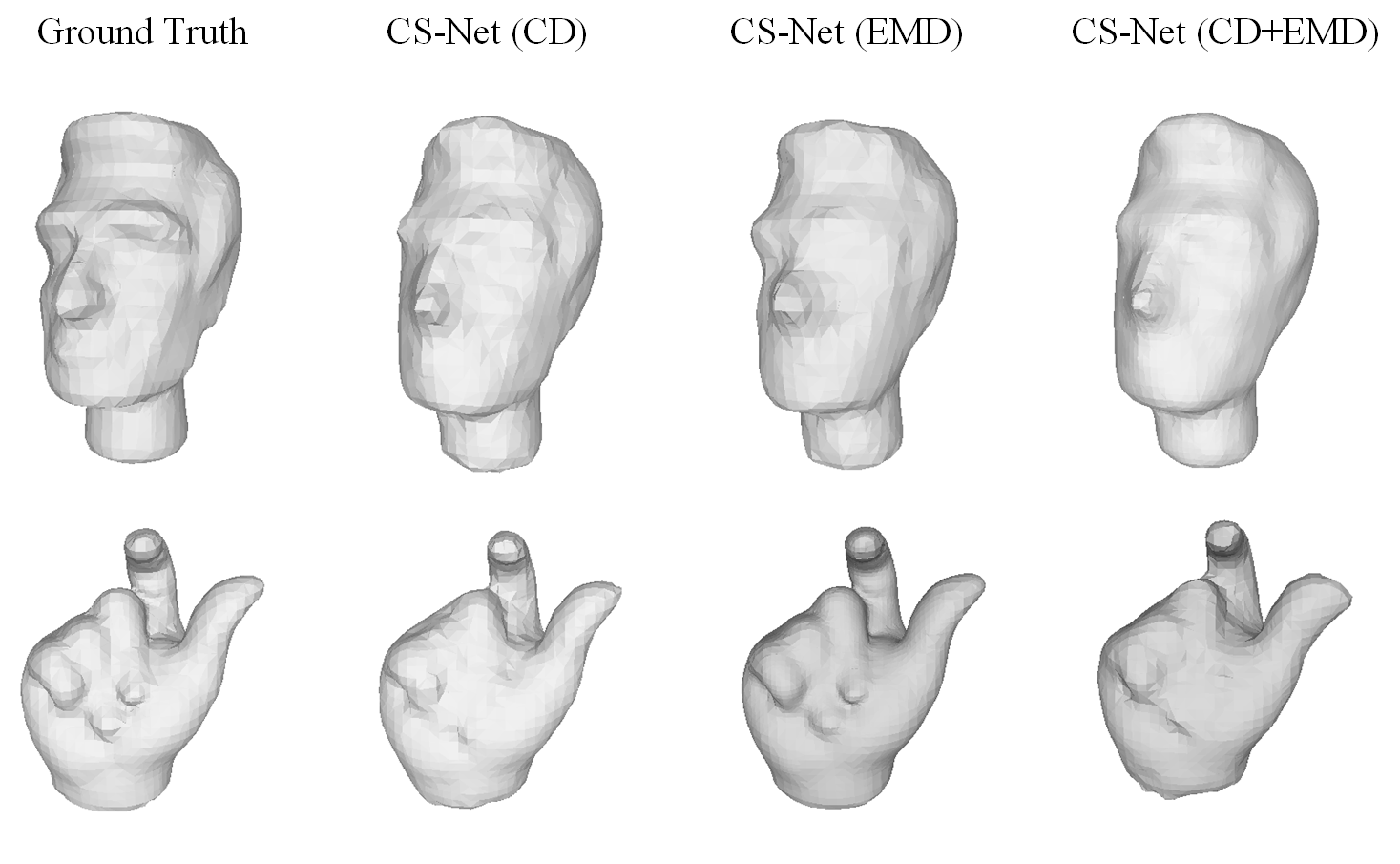}
\caption{Surface reconstruction with different loss functions.}
\label{FIG14}
\end{figure}

Fig. 14 shows that surface reconstruction using EMD was significantly better than reconstruction using CD. Due to the differing optimization objectives of EMD and CD, surface reconstruction using both CD and EMD was not as effective as reconstruction using EMD alone.

\section{Conclusion}
We proposed a point cloud sampling network based on point-wise contribution, in which we addressed the selective sampling problem by a Top-\(k\) operation. In the proposed network, a spatial pooling in the feature embedding module is first introduced to extract both global features and local details. Then, a cascade attention module is proposed to enhance the extracted features. Next, a contribution scoring module is proposed to sample points directly from the input point cloud  by the Top-\(k\) operation. Experimental results show that the proposed network outperforms existing methods in classification, registration, sampling-based compression, and surface reconstruction. In the future, we will further focus on reducing the GPU memory consumption of the proposed method (which is also the common problem of most deep-learning-based sampling methods) and making it suitable to directly handle large-scale point clouds or dense point clouds with millions of points without patch-based processing.

\vfill


\begin{thebibliography}{1}
\bibliographystyle{IEEEtran}

\bibitem{refb}
X. Li, R. Li, G. Chen, C. -W. Fu, D. Cohen-Or, and P. -A. Heng, “A Rotation-Invariant Framework for Deep Point Cloud Analysis,"  \textit{IEEE Trans. Vis. Comput. Graphics}, vol. 28, no. 12, pp. 4503-4514, 2022.

\bibitem{refc}
A. Mao, Z. Du, J. Hou, Y. Duan, Y. -J. Liu, and Y. He, “PU-Flow: A Point Cloud Upsampling Network With Normalizing Flows," \textit{IEEE Trans. Vis. Comput. Graphics}, vol. 29, no. 12, pp. 4964-4977, 2023.

\bibitem{refa}
Y. Wu, \textit{et al.}, “Evolutionary Multiform Optimization With Two-Stage Bidirectional Knowledge Transfer Strategy for Point Cloud Registration," \textit{IEEE Trans. Evol. Comput.}, vol. 28, no. 1, pp. 62-76, 2024.

\bibitem{refd}
Y. Yuan, Y. Wu, X. Fan, M. Gong, W. Ma, and Q. Miao, “EGST: Enhanced Geometric Structure Transformer for Point Cloud Registration," \textit{IEEE Trans. Vis. Comput. Graphics}, early access, 2023, doi: 10.1109/TVCG.2023.3329578.

\bibitem{refd2}
T. Guo, H. Yuan, R. Hamzaoui, X. Wang, and L. Wang, “Dependence-Based Coarse-to-Fine Approach for Reducing Distortion Accumulation in G-PCC Attribute Compression," \textit{IEEE Trans. Ind. Informat.}, early access, 2024, doi: 10.1109/TII.2024.3403262.

\bibitem{ref1}
S. Qiu, S. Anwar, and N. Barnes, “PnP-3D: A Plug-and-Play for 3D Point Clouds," \textit{IEEE Trans. Pattern Anal. and Mach. Intell.}, vol. 45, no. 1, pp. 1312-1319, 2023.

\bibitem{ref2}
S. Qiu, S. Anwar, and N. Barnes, “Geometric Back-Projection Network for Point Cloud Classification," \textit{IEEE Trans. Multimedia}, vol. 24, pp. 1943-1955, 2022.

\bibitem{refq}
C. Saltori, F. Galasso, G. Fiameni, N. Sebe, F. Poiesi, and E. Ricci, “Compositional Semantic Mix for Domain Adaptation in Point Cloud Segmentation," \textit{IEEE Trans. Pattern Anal. and Mach. Intell.}, vol. 45, no. 12, pp. 14234-14247, 2023.

\bibitem{refw}
D. Liu, and W. Hu, “Imperceptible Transfer Attack and Defense on 3D Point Cloud Classification," \textit{IEEE Trans. Pattern Anal. and Mach. Intell.}, vol. 45, no. 4, pp. 4727-4746, 2023.

\bibitem{ref3}
H. Liu, H. Yuan, R. Hamzaoui, W. Gao, and S. Li, “PU-Refiner: A geometry refiner with adversarial learning for point cloud upsampling," in \textit{Proc. IEEE Int. Conf. Acoust., Speech and Signal Process.}, 2022, pp. 2270-2274.

\bibitem{ref4}
Z. Li, G. Li, T. H. Li, S. Liu, and W. Gao, “Semantic Point Cloud Upsampling," \textit{IEEE Trans. Multimedia}, vol. 25, pp. 3432-3442, 2023.

\bibitem{refe}
H. Liu, H. Yuan, J. Hou, R. Hamzaoui, and W. Gao, “PUFA-GAN: A Frequency-Aware Generative Adversarial Network for 3D Point Cloud Upsampling," \textit{IEEE Trans. Image Process.}, vol. 31, pp. 7389-7402, 2022.

\bibitem{refr}
Y. Qian, J. Hou, S. Kwong, and Y. He, “Deep Magnification-Flexible Upsampling Over 3D Point Clouds," \textit{IEEE Trans. Image Process.}, vol. 30, pp. 8354-8367, 2021.

\bibitem{ref5}
C. R. Qi, L. Yi, H. Su, and L. J. Guibas, “PointNet++: Deep hierarchical feature learning on point sets in a metric space,” in \textit{Proc. Adv. Neural Inf. Process. Syst.}, 2017, pp. 5105-5114.

\bibitem{ref6}
X. Ying, S. Xin, Q. Sun, and Y. He, “An intrinsic algorithm for parallel poisson disk sampling on arbitrary surfaces,” \textit{IEEE Trans. Vis. Comput. Graphics}, vol. 19, no. 9, pp. 1425–1437, 2013.

\bibitem{ref9}
O. Dovrat, I. Lang, and S. Avidan, “Learning to sample,” in \textit{Proc. IEEE Conf. Comput. Vis. Pattern Recognit.}, 2019, pp. 2755-2764.

\bibitem{ref10}
I. Lang, A. Manor, and S. Avidan, “SampleNet: Differentiable point cloud sampling,” in \textit{Proc. IEEE Conf. Comput. Vis. Pattern Recognit.}, 2020, pp. 7575-7585.

\bibitem{ref11}
Y. Lin, K. Chen, S. Zhou, Y. Huang, and Y. Lei, “CO-NET: Classification-oriented point cloud sampling via informative feature learning and non-overlapped local adjustment," in \textit{Proc. IEEE Int. Conf. Acoust., Speech and Signal Process.}, 2023, pp. 1-5.

\bibitem{ref33}
C. Chen, H. Yuan, H. Liu, J. Hou, and R. Hamzaoui, “CAS-Net: Cascade Attention-Based Sampling Neural Network for Point Cloud Simplification," in \textit{Proc. IEEE Int. Conf. Multimedia and Expo}, 2023, pp. 1991-1996.

\bibitem{ref12}
Y. Lin, Y. Huang, S. Zhou, M. Jiang, T. Wang, and Y. Lei, “DA-Net: Density adaptive down sampling network for point cloud classification via end-to-end learning,” in \textit{Proc. Int. Conf. Pattern Recognit. and Artif. Intell.}, 2021, pp. 13–18.

\bibitem{ref13}
X. Wang, Y. Jin, Y. Cen, C. Lang, and Y. Li, “PST-NET: Point cloud sampling via point based transformer,” in \textit{Proc. Int. Conf. Image and Graphics}, 2021, pp. 57–69.

\bibitem{ref14}
F. Tian, Y. Song, Z. Jiang, W. Tao, and G. Jiang. “UPSNet: Universal Point Cloud Sampling Network Without Knowing Downstream Tasks,” \textit{Inf. Technol. and Contr.}, vol. 51, no. 4, pp. 723-737, 2022.

\bibitem{ref8}
Q. Hu, Bo Yang, L. Xie, S. Rosa, Y. Guo, Z. Wang, N. Trigoni, and A. Markham, “RandLA-Net: Efficient semantic segmentation of large-scale point clouds,” in \textit{Proc. IEEE Conf. Comput. Vis. Pattern Recognit.}, 2020, pp. 11105-11114. 

\bibitem{ref15}
Y. Qian, J. Hou, Q. Zhang, Y. Zeng, S. Kwong, and Y. He, “Task-oriented compact representation of 3D point clouds via a matrix optimization-driven network," \textit{IEEE Trans. Circuits and Syst. Video Technol.}, vol. 33, no. 11, pp. 6981-6995, 2023.

\bibitem{ref16}
Y. Yang, A. Wang, D. Bu, Z. Feng, and J. Liang, “AS-Net: An attention-aware downsampling network for point clouds oriented to classification tasks,” \textit{ Visual Commun. and Image Represent.}, vol. 89, pp. 103639, 2022.

\bibitem{ref17}
J. Yang, Q. Zhang, B. Ni, L. Li, J. Liu, M. Zhou, and Q. Tian, “Modeling point clouds with self-attention and gumbel subset sampling,” in \textit{Proc. IEEE Conf. Comput. Vis. Pattern Recognit.}, 2019, pp.3323–3332.

\bibitem{ref18}
R. Sun, G. Chen, J. Ma, and P. An, “Straight sampling network for point cloud learning,” in \textit{Proc. IEEE Int. Conf. Image Process.}, 2021, pp. 3088–3092.

\bibitem{ref19}
E. Nezhadarya, E. Taghavi, R. Razani, B. Liu, and J. Luo, “Adaptive hierarchical downsampling for point cloud classification,” in \textit{Proc. IEEE Conf. Comput. Vis. Pattern Recognit.}, 2020, pp. 12953-12961.

\bibitem{refd1}
R.A. Potamias, G. Bouritsas, and S. Zafeiriou, “Revisiting Point Cloud Simplification: A Learnable Feature Preserving Approach," in \textit{Proc. Eur. Conf. Comput. Vis.}, 2022, pp. 586-603.

\bibitem{ref21}
Y. Xie, H. Dai, M. Chen, B. Dai, T. Zhao, H. Zha, W. Wei, and T. Pfister, “Differentiable Top-k Operator with Optimal Transport,” in \textit{Proc. Adv. Neural Inf. Process. Syst.}, 2020, pp. 20520–20531.

\bibitem{ref22}
M. Cuturi, O. Teboul, and J. Vert, “Differentiable ranking and sorting using optimal transport,” in \textit{Proc. Adv. Neural Inf. Process. Syst.}, 2019, pp. 6861–6871.

\bibitem{refcb}
S. Woo, J. Park, JY. Lee, and I.S. Kweon, “CBAM: Convolutional Block Attention Module,” in \textit{Proc. Eur. Conf. Comput. Vis.}, 2018, pp. 3-19.

\bibitem{ref23}
A. Vaswani, N. Shazeer, N. Parmar, J. Uszkoreit, L. Jones, A. N. Gomez, Ł. Kaiser, and I. Polosukhin, “Attention is all you need,” in \textit{Proc. Adv. Neural Inf. Process. Syst.}, 2017, pp. 6000–6010.

\bibitem{ref24}
K. He, X. Zhang, S. Ren, and J. Sun, “Deep residual learning for image recognition,” in \textit{Proc. IEEE Conf. Comput. Vis. Pattern Recognit.}, 2016, pp. 770-778.

\bibitem{ref25}
M. Guo, J. Cai, Z. Liu, T. Mu, R. Martin, and S. Hu, “PCT: Point cloud transformer,” \textit{Comput. Visual Media}, vol. 7, no. 2, pp. 187–199, 2021.

\bibitem{ref32}
H. Ling and K. Okada, “An efficient earth mover's distance algorithm for robust histogram comparison," \textit{IEEE Trans. Pattern Anal. and Mach. Intell.}, vol. 29, no. 5, pp. 840-853, 2007.

\bibitem{ref26}
Z. Wu, S. Song, A. Khosla, F. Yu, L. Zhang, X. Tang, and J. Xiao, “3D ShapeNets: A deep representation for volumetric shapes,” in \textit{Proc. IEEE Conf. Comput. Vis. Pattern Recognit.}, 2015, pp. 1912-1920.

\bibitem{ref29}
R. Li, X. Li, C. -W. Fu, D. Cohen-Or, and P. -A. Heng, “PU-GAN: A Point Cloud Upsampling Adversarial Network," in \textit{Proc. IEEE Int. Conf. Comput. Vis.}, 2019, pp. 7202-7211. 

\bibitem{refkitti}
A. Geiger, P. Lenz, and R. Urtasun, “Are we ready for autonomous driving? The KITTI vision benchmark suite," in \textit{Proc. IEEE Conf. Comput. Vis. Pattern Recognit.}, 2012, pp. 3354-3361.

\bibitem{ref27}
C. Ruizhongtai Qi, H. Su, K. Mo, and L. J. Guibas, “PointNet: Deep learning on point sets for 3d classification and segmentation,” in \textit{Proc. IEEE Conf. Comput. Vis. Pattern Recognit.}, 2017, pp. 77-85.

\bibitem{ref30}
J. Yang, H. Li, D. Campbell, and Y. Jia, “Go-ICP: A Globally Optimal Solution to 3D ICP Point-Set Registration,” \textit{IEEE Trans. Pattern Anal. and Mach. Intell.}, vol. 38, no. 11, pp. 2241-2254, 2016.

\bibitem{ref28}
3DG, “G-PCC Test Model v19,” \textit{Doc. ISO/IEC JTC1/SC29/WG7 MPEG N0360}, 2022.

\bibitem{ref31}
K. Michael, and H. Hoppe, “Screened Poisson surface reconstruction," \textit{ACM Trans. on Graphics}. vol. 32, no. 3, pp. 1-13, 2013.

\bibitem{refvoxel}
J. Deng, S. Shi, P. Li, W. Zhou, Y. Zhang, and H. Li, “Voxel R-CNN: Towards High Performance Voxel-based 3D Object Detection,” in \textit{Proc. AAAI Conf. Artif. Intell.}, 2017, pp. 77-85.

\end{thebibliography}
\end{document}